\documentclass{article}

\usepackage[preprint]{neurips_2026}

\usepackage[T1]{fontenc}
\usepackage[scaled=0.85]{DejaVuSansMono}

\usepackage[utf8]{inputenc} 
\usepackage[T1]{fontenc}    
\usepackage[breaklinks]{hyperref}       
\usepackage{url}            
\usepackage{booktabs}       
\usepackage{amsfonts}       
\usepackage{nicefrac}       
\usepackage{microtype}      
\usepackage{tikz}
\usepackage[table]{xcolor}
\usepackage{lipsum}
\usepackage{tabularx}
\usepackage{array}
\usepackage{url}
\usepackage{graphicx}
\usepackage{wrapfig}
\usepackage{subcaption}
\usepackage{booktabs} 
\usepackage{multirow}
\usepackage{diagbox}
\usepackage{adjustbox}
\usepackage{enumitem}
\usepackage{listings}
\usepackage{pifont}

\usepackage{amsmath,amsfonts,bm}

\def\eqref#1{equation~\ref{#1}}

\def\1{\bm{1}}

\DeclareMathAlphabet{\mathsfit}{\encodingdefault}{\sfdefault}{m}{sl}
\SetMathAlphabet{\mathsfit}{bold}{\encodingdefault}{\sfdefault}{bx}{n}

\definecolor{kellygreen}{rgb}{0.3, 0.73, 0.09}
\definecolor{alizarin}{rgb}{0.82, 0.1, 0.26}
\newcommand{\cmark}{{\color{kellygreen} \ding{51}}}
\newcommand{\xmark}{{\color{alizarin} \ding{55}}}

\title{
KV Cache Offloading for Context-Intensive Tasks
}

\author{%
  \hspace{-30px}Andrey Bocharnikov\thanks{Equal contribution.} \\
  \hspace{-30px}HSE, Yandex
  \And
  \hspace{-10px}Ivan Ermakov\footnotemark[1] \\
  \hspace{-10px}HSE, Yandex
  \And
  \hspace{-10px}Denis Kuznedelev\footnotemark[1]\thanks{Correspondence to: \texttt{dkuznedelev@yandex-team.ru}} \\
  \hspace{-10px}Yandex
  \And
  \hspace{-10px}Vyacheslav Zhdanovskiy\footnotemark[1] \\
  \hspace{-10px}Yandex
  \And
  \hspace{-10px}Yegor Yershov\footnotemark[1]\hspace{-40px} \\
  \hspace{-10px}Yandex, NSU\hspace{-40px}
}

\begin{document}

\maketitle

\begin{abstract}
With the growing demand for long-context LLMs across a wide range of applications, the key–value (KV) cache has become a critical bottleneck for both latency and memory usage. 
Recently, KV-cache offloading has emerged as a promising approach to reduce memory footprint and inference latency while preserving accuracy.
Most recent works on KV offloading evaluate on tasks that do not require extracting large amounts of information from the context.
In this work, we study KV-cache offloading on \underline{context-intensive} tasks: problems where the solution requires looking up a lot of information from the input prompt.
We create and release the Text2JSON benchmark, a highly context-intensive task that requires extracting structured knowledge from raw text.
We evaluate modern KV offloading on Text2JSON and other context-intensive tasks and find significant performance degradation on both Llama 3 and Qwen3 models.
Our analysis identifies two key reasons for poor accuracy: low-rank projection of keys and unreliable selection, and proposes a simpler alternative strategy that significantly improves accuracy across multiple LLM families and benchmarks.
These findings highlight the need for a comprehensive and rigorous evaluation of long-context compression techniques.
\end{abstract}

\vspace{-10px}
\section{Introduction}\label{sect:introduction}
\vspace{-10px}

Large Language Models (LLMs) are becoming increasingly adept at handling long-context inputs, enabling applications such as long-document summarization \cite{bai2024longbench, wang2024leave}, large-scale codebase analysis \cite{liu2023repobench, luo2024repoagent}, and agentic workflows, and scaling to even longer tasks~\cite{measuring-ai-ability-to-complete-long-tasks}. However, processing long contexts remains challenging because the key–value (KV) cache grows linearly with sequence length. For sufficiently long sequences, the KV cache can require more memory than the model parameters themselves \citep{hooper2024kvquant}. Moreover, as the KV cache grows in size, it often becomes the primary bottleneck for LLM inference.
Since KV cache size scales with sequence length, long-context inference can fit fewer sequences in accelerator memory, drastically reducing inference throughput~\cite{zhang2023h2o,sheng2023flexgen}.

The two most popular ways to reduce KV cache footprint are quantization~\citep{hooper2024kvquant, liu2024kivi, shutova2025cache, ashkboos2024quarot} and cache eviction (or pruning)~\citep{xiao2023efficient, zhang2023h2o, li2024snapkv, tang2024quest}.
KV quantization has already gained significant popularity among practitioners~\cite{vllm2026kv,trtllm2026quant}. In turn, KV eviction is not commonly used in industrial deployments~\cite{li2025snapstream}, largely because pruning KV entries can lead to significant performance drops for some tasks~\cite{chen2025pitfalls,ananthanarayanan2026understanding}. For instance, KV eviction struggles with problems that require processing most of the input tokens: sorting a document collection, translating a book verbatim or refactoring a codebase.

A promising alternative to pruning is KV-cache offloading~\citep{aminabadi2022deepspeed, sheng2023flexgen}: instead of removing KV entries permanently, these techniques move the original entries to cheaper system memory and load them back when necessary. To avoid loading the full KV cache, they use method-specific heuristics to predict which KV entries will be used at every inference step~\cite{lee2024infinigen,chen2024arkvale, sun2025shadowkv,tenghui2026efficient,yang2025hcattentionextremekvcache}, reporting little to no performance degradation compared to full attention. However, evaluations in these studies have largely been restricted to widely used long-context benchmarks such as RULER~\citep{hsieh2024ruler} and LongBench \citep{bai2024longbench}. Though popular, these benchmarks may not capture the full range of real-world applications. As a result, performance drops on practically relevant tasks may go unnoticed and could be non-negligible.\nocite{li2024looglelongcontextlanguagemodels}

In this work, we systematically evaluate KV-cache offloading techniques across more difficult workloads and identify one prominent failure mode: \textit{context-intensive tasks}.
Namely, we find that KV offloading methods struggle with problems that require looking up a lot of contextual information, even if the problem itself is not difficult. We identify several context-intensive subtasks in existing benchmarks that are not common in KV offloading research. Additionally, we create a new dataset, Text2JSON, based on the real-world task of extracting structured information from documents.

We systematically evaluate KV offloading methods on context-intensive tasks and observe significant accuracy drops. Further analysis attributes this to three main problems: 1) suboptimal key compression, 2) inaccurate selection criteria and 3) key/value grouping. Curiously, these components do not hinder accuracy on needle-in-a-haystack tasks, but become problematic when the LLM needs to retrieve much more information.
Our main contributions can be summarized as follows:\begin{enumerate}[leftmargin=*]
    \vspace{-5px}\item We identify context-intensive subtasks in existing benchmarks and create a new Text2JSON dataset to evaluate KV offloading on real-world context-intensive tasks without confounding factors.
    \vspace{-3px}\item We evaluate several modern KV offloading methods across two LLM families (Llama and Qwen3) and find consistent accuracy drops on context-intensive problems.
    \vspace{-3px}\item We analyze the failure modes of KV offloading and find that the accuracy loss is caused by key selection strategies, not KV offloading in principle. Crucially, the same optimizations do not harm accuracy on non-intensive tasks such as RULER, highlighting a gap for future evaluations.
    \vspace{-3px}\item Using our findings, we formulate Yet Another KV Offloading\footnote{YAKV Offloading} algorithm that replaces the problematic components with simple data-free quantization. The resulting algorithm achieves near-lossless accuracy on context-intensive tasks while maintaining high inference throughput.
\end{enumerate}\vspace{-3px}



\vspace{-10px}
\section{Background}\label{sect:background}
\vspace{-10px}

Transformer-based LLMs typically store task-specific information in a key-value (KV) cache. This cache contains token-level vector representations that are consumed by the attention layers at each inference step~\citep{vaswani2017attention}. In modern LLMs, the cache is usually limited to ${\approx} 10^{4-6}$ tokens, with more extreme cases exceeding one million tokens~\citep{glm2024chatglmfamilylargelanguage,yang2025qwen251mtechnicalreport}. Though this limit can be extended~\citep{yarn,longchat,pekelis2024llama3gradient}, large KV caches quickly exhaust accelerator memory\footnote{For example, storing 1M tokens for \href{https://huggingface.co/Qwen/Qwen2.5-7B-Instruct-1M}{Qwen2.5-7B-Instruct-1M} in \texttt{bfloat16} requires over 180\,GiB of GPU memory.} and reduce inference throughput.

To mitigate this, several lines of work propose KV-cache quantization~\citep{liu2024kivi,hooper2024kvquant,ashkboos2024quarot}, eviction (pruning)~\citep{xiao2023efficient,zhang2023h2o,li2024snapkv}, cross-layer sharing~\cite{wu2025systematicstudycrosslayerkv,yang2024kvsharer,lin2026reconstructing}, and other related techniques. These approaches are not mutually exclusive: an inference server may evict a subset of KV entries and quantize the remainder to further reduce memory usage~\cite{li2024snapkv,aquakv}. However, while quantization is already widely used in industrial deployments~\cite{vllm_quantized_kv_cache,tensorrt_llm_quantization_blog_including_kv}, KV eviction has seen slower adoption. A key reason is that pruning KV entries can significantly degrade performance on certain problem types~\cite{li2025snapstreamefficientlongsequence,chen2025pitfallskvcachecompression,ananthanarayanan2026understandingphysicskeyvaluecache} where it is hard to guess which entries can be evicted, or if the problem needs the entire prompt (e.g. translation).

A more recent line of work proposes an alternative: instead of permanently removing KV-cache entries, these methods offload KV vectors to a larger but slower system memory~\citep{aminabadi2022deepspeed, sheng2023flexgen,mohtashami2023landmark}. To avoid reloading the entire cache, they estimate which tokens are most relevant at a given step~\citep{lee2024infinigen, chen2024arkvale, sun2025shadowkv, liu2024retrievalattention, mohtashami2023landmark}, using mechanisms closely related to sparse attention~\citep{tang2024quest, liu2021transformer, yuan2025native, jiang2024minference}. For instance, InfiniGen~\cite{lee2024infinigen} uses SVD approximation to determine important keys and load them from system memory concurrently with previous layer computation. ShadowKV~\cite{sun2025shadowkv} further splits KV entries into groups of 8 consecutive tokens and uses group-average keys (``landmarks'') to determine which KV entries should be loaded at a given inference step. Subsequent works extend this idea further with better ``landmarks''~\cite{chen2024arkvale}, specialized key decomposition~\cite{tenghui2026efficient} and technical optimizations~\cite{yang2025hcattentionextremekvcache}.
Another line of work ~\cite{magicpig,pqcache} proposes low-bit representations
for important token identification.
Unlike permanent pruning, offloading preserves the possibility of accessing the full KV cache over time, and is therefore, in principle, capable of solving problems that pruning would struggle with. In this work, we find that this promise does not always hold in practice and propose a way to fix it.

\vspace{-10px}
\section{Context-Intensive Tasks}\label{sect:method}
\vspace{-10px}

The core intuition of our work is that long-context tasks vary in how much information they need to extract in order to solve the problem correctly. On one end of the spectrum Needle in the Haystack (NIAH)~\cite{hsieh2024ruler} requires the model to look up a single correct substring (needle) in a long prompt. On the other end, there are real-world problems that require looking up and cross-referencing many interdependent ``needles''. Such context-intensive tasks include context-aware document translation, programming in a pre-existing codebase, knowledge extraction, legal case analysis, and others.

\begin{table*}[t]
\centering
\caption{Long-context benchmarks in terms of context intensity, scenario type and use in KV eviction, selection and offloading evaluation; \cmark/\xmark\, means that only some subtasks of this benchmark apply.}
\label{tab:benchmark_comparison}
\small
\resizebox{0.8\linewidth}{!}{
\begin{tabular}{l c c c c c}
\toprule
Benchmark & Context   & Realistic & Used in  & Used in  & Used in  \\
Title     & Intensity & Scenario  & Cache Eviction & Sparse Attention & KV Offloading \\
\midrule
NIAH~\cite{kamradt2023needle} & Low & \xmark & \cmark & \cmark & \cmark \\
RULER~\cite{hsieh2024ruler} & Low & \xmark & \cmark & \cmark & \cmark \\
Scrolls~\cite{shaham-etal-2022-scrolls} & Low & \cmark & \cmark & \cmark & \cmark / \xmark \\
LongBench~\cite{bai2024longbench,bai2024longbench2} & Low & \cmark / \xmark & \cmark / \xmark & \cmark / \xmark & \cmark / \xmark \\
SCBench~\cite{li2024scbench} & Low & \cmark / \xmark & \cmark & \cmark & \xmark \\
$\infty$Bench~\cite{zhang-etal-2024-infinitebench} & Low & \cmark & \cmark & \cmark & \xmark \\
BAMBOO~\cite{dong2024bamboo} & Low & \cmark / \xmark & \xmark & \xmark & \xmark \\
L-eval~\cite{an2024eval} & Low & \cmark & \xmark & \xmark & \xmark \\
BABILong~\cite{kuratov2024babilong} & Medium & \xmark & \xmark & \xmark & \xmark \\
Marathon~\cite{zhang2024marathon} & Medium & \cmark & \xmark & \xmark & \xmark \\
LooGLE v2~\cite{li2024looglelongcontextlanguagemodels,he2025looglev2llmsready} & Medium & \cmark & \xmark & \xmark & \xmark \\
NeedleBench v2~\cite{li2025needlebench} & High & \xmark & \xmark & \xmark & \xmark \\
LongProc~\cite{ye25longproc} & High & \xmark & \xmark & \xmark & \xmark \\
Loong~\cite{wang2024leave} & High & \cmark & \xmark & \xmark & \xmark \\
\bottomrule
\end{tabular}
}
\vspace{-15px}
\end{table*}

Most popular benchmarks used to evaluate KV offloading are not context-intensive. The RULER benchmark~\cite{hsieh2024ruler}, for example, consists of synthetic needle-in-a-haystack~\cite{kamradt2023needle} tasks where the model only needs to extract a single (or very few) text ``needles'' from long context. More advanced NIAH benchmarks~\cite{bianchi2025SmallerNeedles,chen2021finqa,hengle-etal-2025-multilingual,longmemeval,yuan2025lveval} introduce more challenging domains or distractors to make finding the correct ``needle'' harder; however, the total amount of information required to solve each problem remains small. Similarly, LongBench~\cite{bai2024longbench} aggregates 21 tasks with heterogeneous semantics, including single- and multi-document question answering, synthetic counting tasks, and code completion. Only a small subset of these tasks is relatively context-intensive (e.g., passage\_count), and these tasks are both underrepresented and often excluded from evaluation protocols~\cite{chen2024arkvale,sun2025shadowkv}.

We collate long-context benchmarks in Table~\ref{tab:benchmark_comparison}. For simplicity, we consider a benchmark highly context-intensive if it requires at least  10 ``needles'' (substrings) to solve the problem correctly, on average using the same aggregation as the benchmark itself. Conversely, 3-10 needles are medium context intensity and less than 3 needles (e.g. NIAH) are low context intensity. For each benchmark, we check if it is used for evaluating post-training KV eviction~\cite{zhang2023h2o,xiao2023efficient,cai2024pyramidkv,li2024snapkv,kim2025kvzipqueryagnostickvcache}\nocite{jegou2026kvzap}, sparse attention~\cite{jiang2024minference,tang2024quest,singhania2406loki,lin2025twilightadaptiveattentionsparsity,zhang2025spargeattention}, and KV offloading~\cite{lee2024infinigen,chen2024arkvale,sun2025shadowkv,tenghui2026efficient,yang2025hcattentionextremekvcache}. To summarize, \textbf{post-training KV offloading methods evaluate almost exclusively on low context-intensity benchmarks}.

Higher context-intensity datasets do exist, typically as subsets of larger benchmarks~\cite{li2025needlebench,wang2024leave,wei2025browsecompsimplechallengingbenchmark,ye25longproc}, but they are not commonly used in KV offloading literature. We list several such benchmarks below:\begin{itemize}[leftmargin=*]
    \vspace{-5px}\item \textbf{MultiNeedle 128K:} NeedleBench v2~\cite{li2025needlebench} includes a subset of synthetic retrieval tasks involving multiple independent ``needles'' or a connected dependency structure (e.g., a family tree). These tasks are simple and artificial, but they are more context-intensive than most NIAH benchmarks~\cite{kamradt2023needle,hsieh2024ruler}. Despite its simplicity, many offloading methods struggle with MultiNeedle (see Appendix~\ref{app:preliminary_experiments}).
    \vspace{-3px}\item \textbf{LongProc~\cite{ye25longproc} HTML to TSV 8K} is a subset of 120 synthetic problems where the LLM must convert structured HTML into a TSV table. These tasks were procedurally generated using websites sourced from Arborist~\cite{li2024arborist} and can largely be solved with a rules-based HTML parser, but still prove counterintuitively difficult for LLMs with KV offloading.
    \vspace{-3px}\item \textbf{Loong}~\cite{wang2024leave} contains multi-document question-answering tasks where each document contributes relevant information needed to produce the final answer, making them more context-intensive than traditional document QA tasks~\cite{bai2024longbench}. Unlike MultiNeedle, Loong offers more realistic evaluation scenarios. However, for our purposes, Loong conflates context intensity with special domain knowledge and multilinguality. Additionally, it uses costly LLM-as-a-Judge~\cite{zheng2023llm-as-a-judge} evaluation. For this reason, we experiment with other benchmarks first and use Loong for verification.    
\end{itemize}\vspace{-3px}

Note also that these benchmarks were not built for evaluating KV offloading or eviction: NeedleBench v2 is a NIAH-type benchmark for general LLM capabilities; LongProc is a long procedural generation benchmark where only 1 out of 17 tasks has inputs longer than 16K tokens on average. On the other hand, Loong contains real-world multi-document QA tasks that are highly context-intensive, but they require domain-specific scientific and financial knowledge, English-Chinese multilinguality and reasoning. This makes it difficult to decouple the effects of KV offloading. To address these limitations, we collect a new Text2JSON dataset to augment these three in our analysis.

\begin{figure}[t]
    \centering
    \vspace{-25px}
    \begin{tikzpicture}
    \node[inner sep=0] (img) {
        \includegraphics[width=0.99\linewidth]{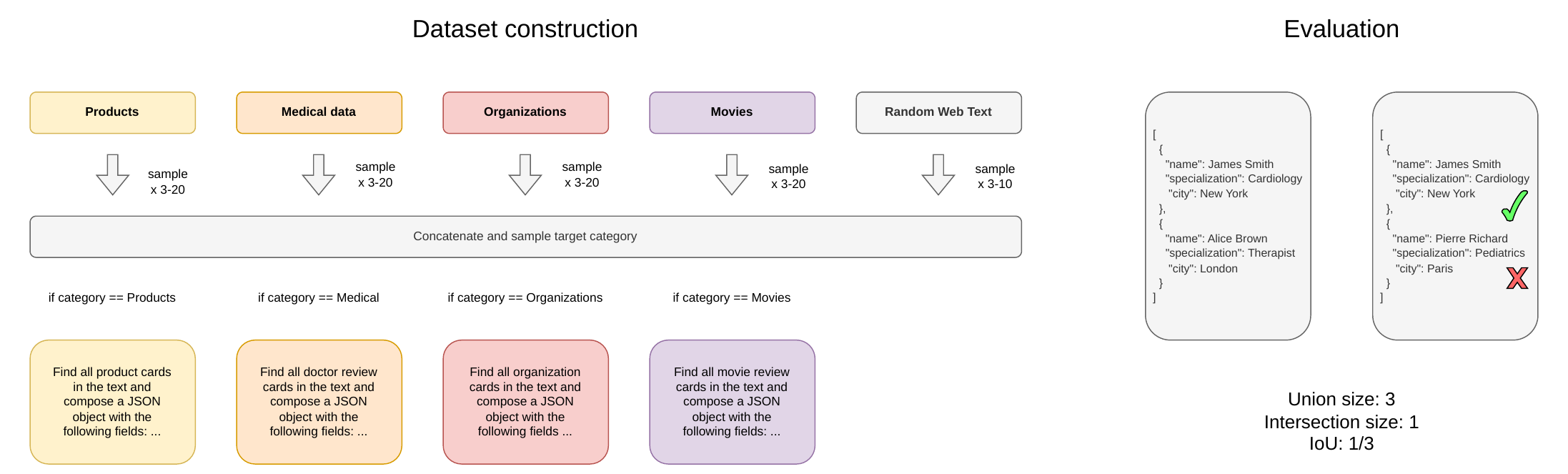}
    };
    \end{tikzpicture}
    \caption{A high-level summary of Text2JSON dataset composition and evaluation.}
    \label{fig:teaser}
    \vspace{-15px}
\end{figure}

\vspace{-7px}
\subsection{Text2JSON: Multiple Needles in the Wild}\label{sect:method_text2json}
\vspace{-7px}

To decouple context-intensity from other LLM capabilities, we gather a new dataset of context-intensive problems inspired by a popular production use case: extracting structured knowledge from unstructured text~\cite{yates-etal-2007-textrunner,ReVerb2011,cui-etal-2018-neural} using LLMs~\cite{josifoski-etal-2022-genie,Dagdelen2024,Xu2024}.
This covers a broad range of applications in search engines, chat personalization, social media, legal document analysis, and others.
While not the most ``exciting'' use case, knowledge extraction has seen wide practical use~\cite{goel_langextract_2026,instructor_567labs_2026,shcherbak_contextgem_2025}.
For our Text2JSON benchmark, we gather 500 samples from across 4 data extraction tasks: product specifications, medical professionals, organization records and movies.

To preserve privacy, the real user data is replaced with LLM-generated facts that  mimic real-world structure (with privacy moderation).
The problems vary between 10.0K and 63.5K tokens (gpt2) with an average length of 20.1K. We deliberately chose long but not extreme context lengths so we decouple context intensity from raw length.
The ground-truth JSON answer contains 3-20 entries (info dictionaries), each with multiple keys and values.
To facilitate reproducibility, we avoid LLM-as-a-Judge~\cite{zheng2023llm-as-a-judge}: instead, we compare each individual item via exact match and report IoU (Intersection over Union) accuracy. 
Additional details on dataset curation and evaluation are in Appendix~\ref{app:text2json_construction}.

\vspace{-8px}
\subsection{Takeaways and Yet Another KV Offloading}\label{sect:method_yakv_takeaways}
\vspace{-7px}

In this work, we systematically analyze KV offloading algorithms on Text2JSON as well as three other context-intensive subtasks from above: MultiNeedle (from NeedleBench v2~\cite{li2025needlebench}), HTML to TSV 8K (from LongProc~\cite{ye25longproc}) and the Loong benchmark~\cite{wang2024leave}. To make our approach easier to follow, we summarize our main takeaways here and provide detailed results in Sections~\ref{sect:experiments_svd} through~\ref{sect:experiments_maintable}.

\textbf{Takeaway A: previously sufficient key compression becomes inaccurate.} Modern KV offloading methods use low-rank key compression. ShadowKV~\cite{sun2025shadowkv} compresses attention keys using truncated SVD (recommended rank 160 layer-wide), which was sufficient for low context-intensity tasks they evaluate on.
However, when this is applied to context-intensive tasks, we found that the LLM can no longer reliably select which information it needs, causing significant accuracy drop (see Section~\ref{sect:experiments_svd}).
In turn, InfiniGen~\cite{lee2024infinigen} and LRQK~\cite{tenghui2026efficient} keep the original keys, but propose specialized decompositions to select which keys are loaded: this leads to missing the necessary keys on context-intensive tasks.

\textbf{Takeaway B: Inaccurate group landmarks.} ShadowKV~\cite{sun2025shadowkv} and ArkVale~\cite{chen2024arkvale} split KV vectors into groups of 8-32 consecutive tokens, similarly to how it is done in GPU sparse attention. Each group is represented with a ``landmark'' --- either the channel-wise average of keys in ShadowKV or a cuboid digest approximation in ArkVale. During inference, the algorithm keeps landmarks on GPU and uses them to determine which groups should be loaded from RAM. Similarly to above, this approximation proved sufficient for NIAH-like tasks. However, as we analyze higher context-intensity tasks, this strategy caused many false positives, resulting in unnecessary token loads and reduced accuracy.

\textbf{YAKV Offloading.} To evaluate the practical impact of our analysis, we combine the two takeaways into a simple KV offloading algorithm. YAKV starts from ShadowKV~\cite{sun2025shadowkv} and simplifies it:\begin{itemize}[leftmargin=*]
    \vspace{-5px}\item \textbf{No key SVD:} we replace key SVD with data-free HIGGS quantization~\cite{malinovskii2024pushing} (4-bit, $d{=}2$,$n{=}256$).
    \vspace{-3px}\item \textbf{No grouping / landmarks:} we avoid key groups altogether. Instead, we keep extremely quantized key approximations (HIGGS 1-2bit) on GPU to select which KV pairs are loaded from RAM.
    \vspace{-3px}\item \textbf{No outliers:} YAKV does not need to store ShadowKV-like outliers since it has no landmarks.
    \vspace{-3px}\item \textbf{Streamlined offloading:} we offload both keys and values into system memory without prefetching.
\end{itemize}\vspace{-5px}

YAKV is designed as a simple baseline that combines takeaways A and B with popular industry practices (e.g. quantization instead of SVD). We deliberately keep YAKV minimal to demonstrate that simple KV offloading with our takeaways outperforms more sophisticated methods on context-intensive tasks. That said, we offer simple strategies to further improve it in Appendices~\ref{app:residual_quantization}~\&~\ref{app:top_p_and_top_kp}. 

\pagebreak

\section{Analysis \& Evaluation}\label{sect:experiments}
\vspace{-5px}

Modern KV offloading systems~\cite{lee2024infinigen,sun2025shadowkv,chen2024arkvale,tenghui2026efficient} have multiple interdependent components: landmarks, compression, decomposition, separate buffers for outliers and recent tokens, prefetching, and others. Furthermore, their behavior can vary between model and task pairs. To keep our results organized, we analyze individual components in Sections~\ref{sect:experiments_svd}---\ref{sect:experiments_better_landmarks}, then generalize to all algorithm-benchmark combinations in Section~\ref{sect:experiments_maintable} and evaluate practical inference throughput in Section~\ref{sect:experiments_inference}. 


\begin{figure}[t]
    \centering
    \vspace{-15px}
    \hspace{-12px}\includegraphics[width=0.51\linewidth]{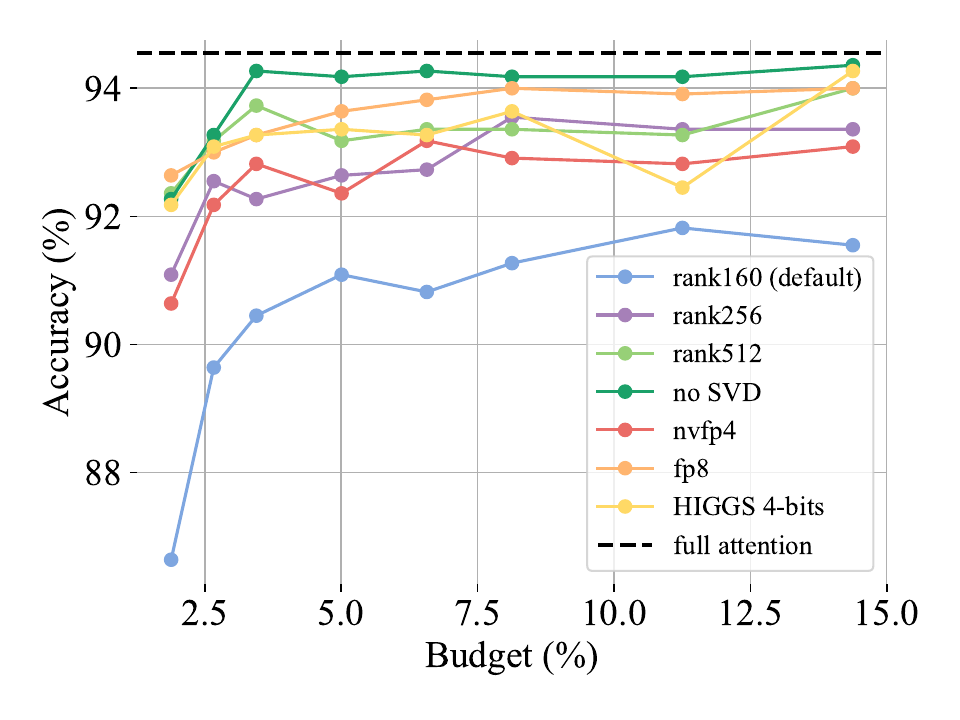}
    \includegraphics[width=0.51\linewidth]{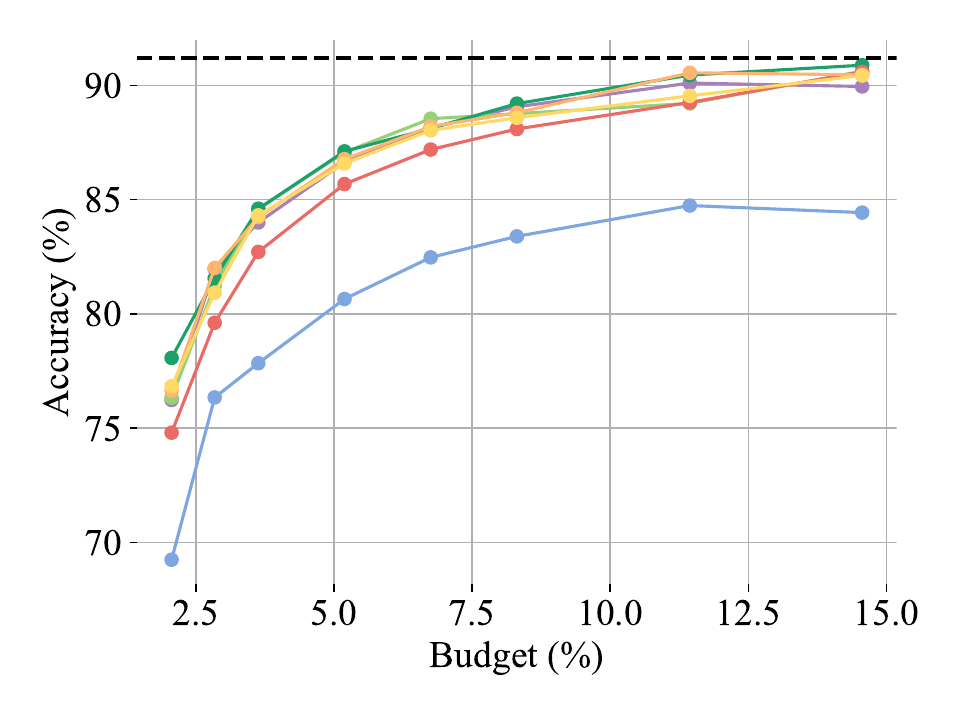}\hspace{-10px}
    
    \vspace{-10px}
    \caption{Evaluation of ShadowKV offloading with different KV compression strategies for Llama-3.1-8B-Instruct on MultiNeedle (left) and Qwen3-4B-Instruct-2507 on Text2JSON (right). The X axis denotes the total percentage of tokens loaded (sparse, outlier and local tokens), see Section~\ref{sect:experiments_svd}.}
    \label{fig:exp_4.1_svd_quant}
    \vspace{-15px}
\end{figure}

\vspace{-5px}
\subsection{Analyzing Key Compression \& Alternatives}\label{sect:experiments_svd}
\vspace{-5px}

KV offloading algorithms often include key and/or value compression to speed up host-to-device transfer~\cite{sheng2023flexgen,sun2025shadowkv,yang2025hcattentionextremekvcache}. Notably, ShadowKV uses an SVD-based decomposition to compress attention keys across heads within one layer \& token, which leads to almost no accuracy degradation on low context-intensity tasks~\cite{sun2025shadowkv}.
Our preliminary experiments (Appendix~\ref{app:preliminary_experiments}) suggest that this compression degrades performance on context-intensive tasks. To analyze this further, we evaluate two models from popular families: Llama-3.1-8B-Instruct~\cite{dubey2024llama} and Qwen3-4B-Instruct-2507~\cite{yang2025qwen3technicalreport}, with additional models in Appendix~\ref{app:experiments_svd}. We use two datasets: the established NeedleBench V2 benchmark~\cite{li2025needlebench} in the MultiNeedle Retrieval 128K setting, and our Text2JSON dataset introduced in Section~\ref{sect:method_text2json}. We use the following KV-cache compression schemes on top of the official ShadowKV code\footnote{\url{https://github.com/ByteDance-Seed/ShadowKV}}:
\begin{itemize}[leftmargin=*]
    \vspace{-5px}\item \textbf{Truncated SVD:} the original ShadowKV uses rank \underline{160}; we additionally report ranks 256 \& 512.
    \vspace{-3px}\item \textbf{Uncompressed:} removing key compression from ShadowKV, at the cost of slower inference.
    \vspace{-3px}\item \textbf{Quantization:} replacing SVD with KV-cache quantization using FP8, NVFP4, or HIGGS-4bit.
\end{itemize}\vspace{-5px}

For quantization, we consider three popular KV-cache compression schemes. FP8 and NVFP4 are compute-oriented quantization formats that are already widely used in deep learning~\cite{micikevicius2022fp8formatsdeeplearning}\nocite{nvidia2022fp8blog} and have also been applied to KV-cache compression~\cite{qiao-etal-2025-swiftkv,nvidia2025nvfp4_kv}. For FP8, we use the E4M3 format. For NVFP4, we follow the protocol of~\cite{egiazarian2026bridginggappromiseperformance}; note that NVFP4 uses micro-scales and averages 4.5 bits per value. HIGGS, by contrast, is a memory-oriented scheme~\cite{malinovskii2024pushing,aquakv} that combines vector quantization with random Hadamard transform; we use a grid with $d=2$ and $n=256$, which averages 4.02 bits per value. Note that certain methods (e.g. LRQK~\cite{tenghui2026efficient}) use alternative key decompositions for key selection, but keep the original keys for evaluation. We analyze these selection methods in Section~\ref{sect:experiments_better_landmarks}.

The results in Figure~\ref{fig:exp_4.1_svd_quant} clearly show that \textbf{the default SVD setting is insufficient} to match the performance of full attention, even when we load $10{\times}$ as many tokens. We attribute this to the greater retrieval difficulty of our context-intensive tasks, where the model must repeatedly attend to correct tokens from the prompt. In this regime, coarse key compression is more likely to introduce retrieval errors. This is not a fundamental limitation of SVD: higher ranks (e.g., 512) perform better. However, at that point, compressing $8{\times}$ 128-dimensional keys (1024 total dimensions) using two projection matrices yields a worse compression ratio than FP8. HIGGS provides a better memory--accuracy trade-off than low-rank SVD, further suggesting that quantization is better suited than aggressive low-rank compression for preserving retrieval quality in these tasks.
We provide additional evaluations in Appendix~\ref{app:experiments_svd}. Based on this observation, we disable SVD in subsequent analysis (Sections~\ref{sect:experiments_budgets_groups}---\ref{sect:experiments_better_landmarks}).

\pagebreak

\vspace{-5px}
\subsection{Analyzing Key Grouping, Landmarks \& Budgets}\label{sect:experiments_budgets_groups}
\vspace{-5px}
\begin{figure}[t]
    \centering
    \vspace{-25px}
    \hspace{-12px}
    \begin{tikzpicture}\node[inner sep=0] (img) {\includegraphics[width=0.51\linewidth]{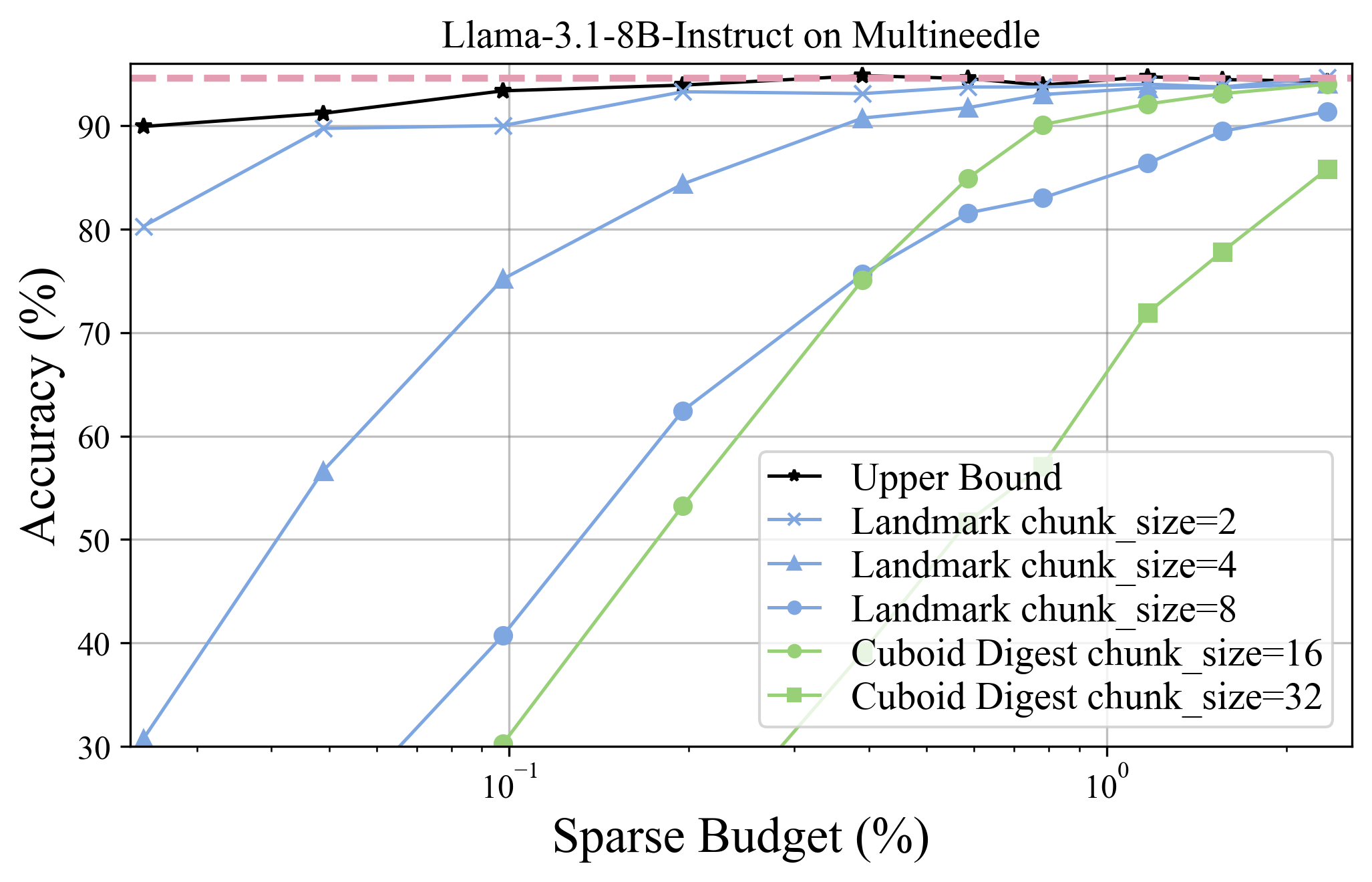}
    };
    \end{tikzpicture}
    \begin{tikzpicture}\node[inner sep=0] (img) {\includegraphics[width=0.51\linewidth]{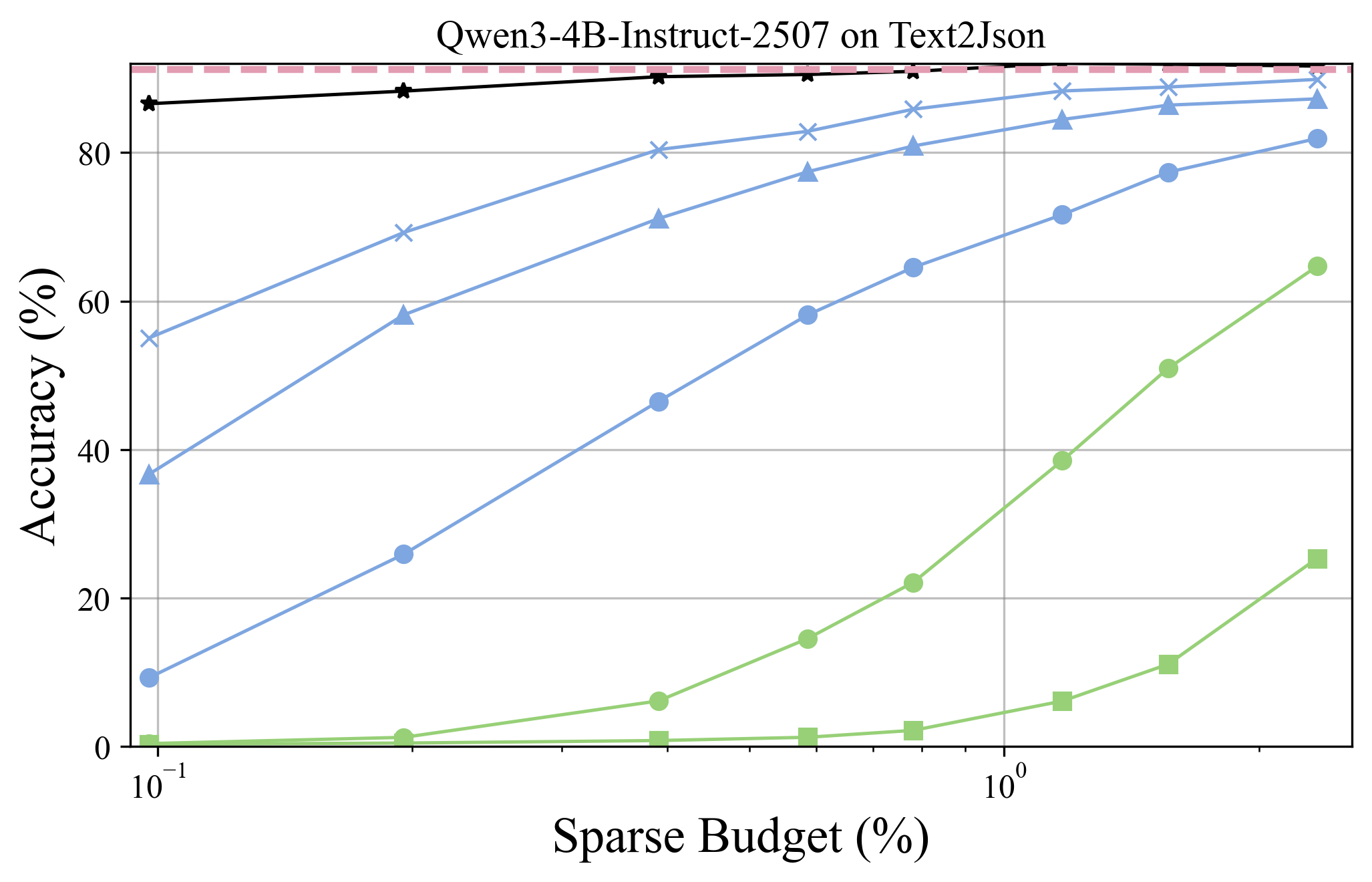}
    };
    \end{tikzpicture}
    \hspace{-10px}
    
    \vspace{-10px}
    \caption{Evaluation of ShadowKV (without SVD) landmarks and ArkVale digests for Section~\ref{sect:experiments_budgets_groups}: \textbf{(left)} Llama-3.1-8B-Instruct on MultiNeedle-128K, and \textbf{(right)} Qwen3-4B-Instruct-2507 on Text2JSON.}
    \label{fig:exp_4.2_grouping}
    \vspace{-15px}
\end{figure}

\begin{figure}[b]
    \centering
    \vspace{-22px}
    \hspace{-12px}
    \includegraphics[width=0.51\linewidth]{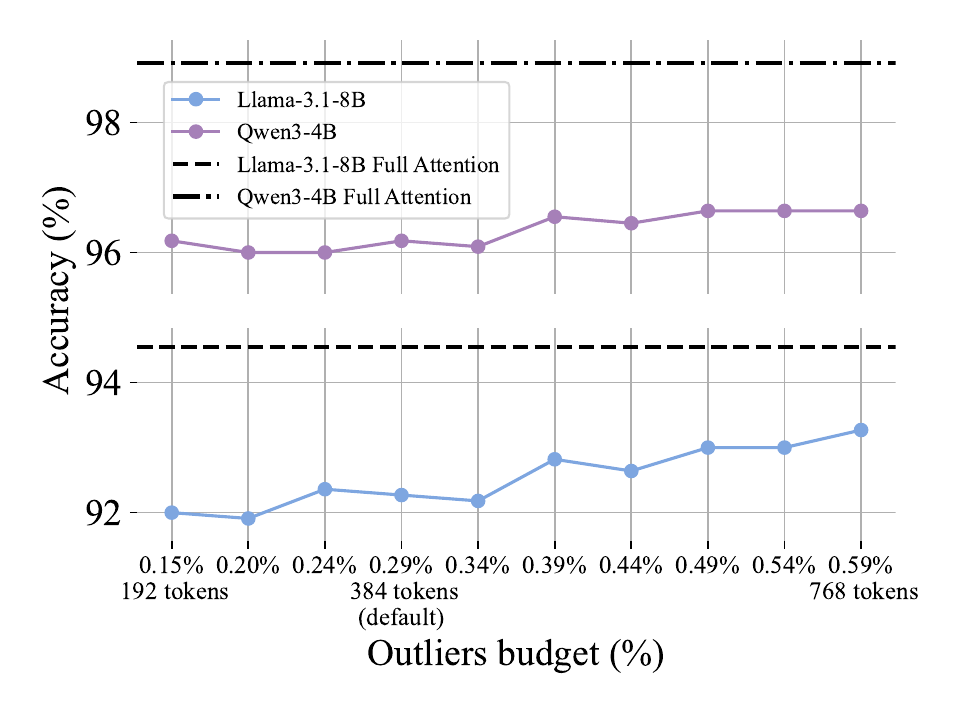}
    \includegraphics[width=0.51\linewidth]{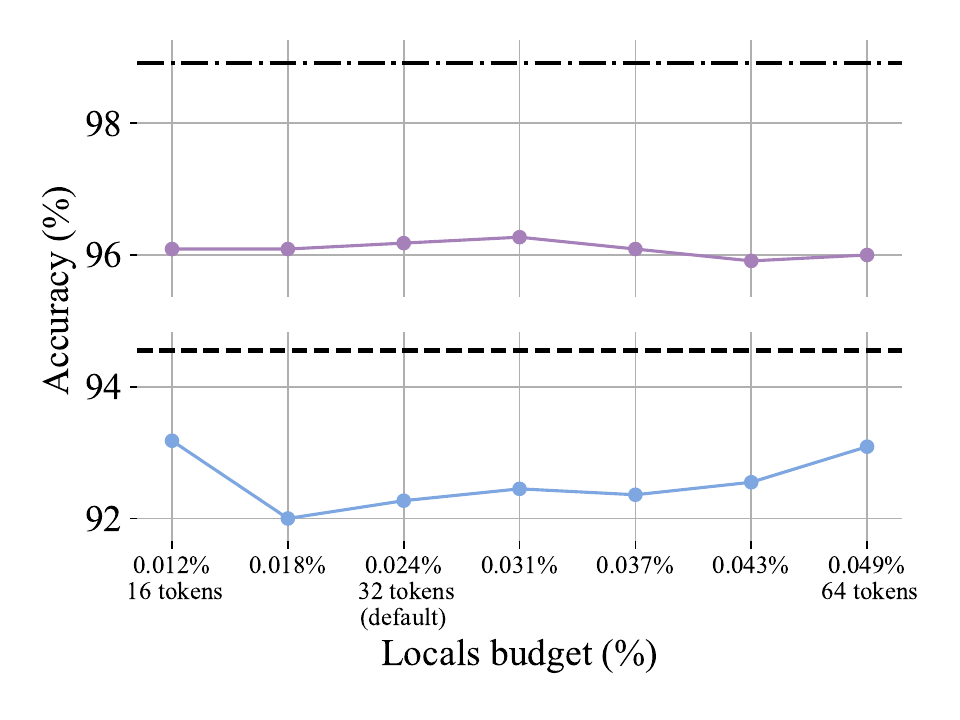}
    \hspace{-10px}
    
    \vspace{-15px}
    \caption{Evaluation of ShadowKV (w/o SVD compression) offloading for Section~\ref{sect:experiments_budgets_groups} with varying outlier budget (left) and local window (right) for Llama-3.1-8B-Instruct and Qwen3-4B-Instruct-2507 on MultiNeedle Retrieval 128K dataset from NeedleBench v2, all other parameters set to default.}
    \label{fig:exp_4.2_local_sparse_budgets}
    \vspace{-10px}
\end{figure}

Without SVD compression, the models can achieve near-lossless accuracy after loading enough tokens from system memory. However, they need to load significantly more tokens to match full attention accuracy: 5-10\% in Figure~\ref{fig:exp_4.1_svd_quant}, compared to the ShadowKV default budget of 1.56\%~\cite{sun2025shadowkv}.
There are two possible explanations: either \textbf{A)} the model fundamentally needs to load more tokens per query or \textbf{B)} the offloading algorithms cannot determine which KVs need to be loaded.

To determine which is the case, we analyze how existing methods select KVs for loading and compare them against an ``oracle'' that always selects the keys with the highest true dot product. Note that the ``oracle'' is not an efficient KV selection algorithm: we use it exclusively as an upper bound on KV selection.
Both ShadowKV and ArkVale use a similar procedure to select which KVs are loaded from RAM: they segment their KV entries into groups (or pages) and compute one ``landmark'' (or digest) per group to determine when it is needed. ShadowKV uses groups of 8 consecutive tokens and summarizes them with a ``landmark'' --- a channel-wise average of keys in that chunk, stored on GPU. The algorithm then computes the dot product between queries and landmarks and loads the highest scoring chunks to GPU. ArkVale uses a slightly more complex bounding-cuboid approximation (digest) that captures how much the keys differ in each consecutive group.

We compare KV selection landmarks from ShadowKV (default group size 8) and ArkVale (default group size 16-32) on top of the same configuration as in Section~\ref{sect:experiments_svd}. The results in Figure~\ref{fig:exp_4.2_grouping} demonstrate that both types of landmarks are inaccurate compared to the ``oracle'' KV selection and need significantly 
more KVs to match full attention accuracy, even with reduced group sizes (2-4).
Similarly to Section~\ref{sect:experiments_svd}, this problem does not affect low context-intensity tasks nearly as much.

To better isolate the source of this inaccuracy, we also test two related components: group outliers and local buffer. In ShadowKV, tokens that do not fit their groups well are called ``outliers'' stored separately (e.g. attention sinks~\cite{xiao2023efficient}). Additionally, most KV offloading methods also treat several most recent tokens this way (``local budget''). In Figure~\ref{fig:exp_4.2_local_sparse_budgets}, we test if increasing either of those budgets can make up for landmark-based KV selection and find that they cannot: even doubling the number of outliers or the local budget (stored on GPU) leaves a significant performance gap compared to full attention. We verify this for additional model-dataset pairs in Appendix~\ref{app:experiments_budgets_groups}.
In summary, \textbf{group-based KV selection is inaccurate on context-intensive tasks}, and this is not easily mitigated with outliers.

\pagebreak
\vspace{-8px}
\subsection{Analyzing \& Improving KV Selection Without Grouping}\label{sect:experiments_better_landmarks}
\vspace{-7px}

\begin{figure}[t]
    \vspace{-25px}
    \centering
    \hspace{-12px}\includegraphics[width=0.5\linewidth]{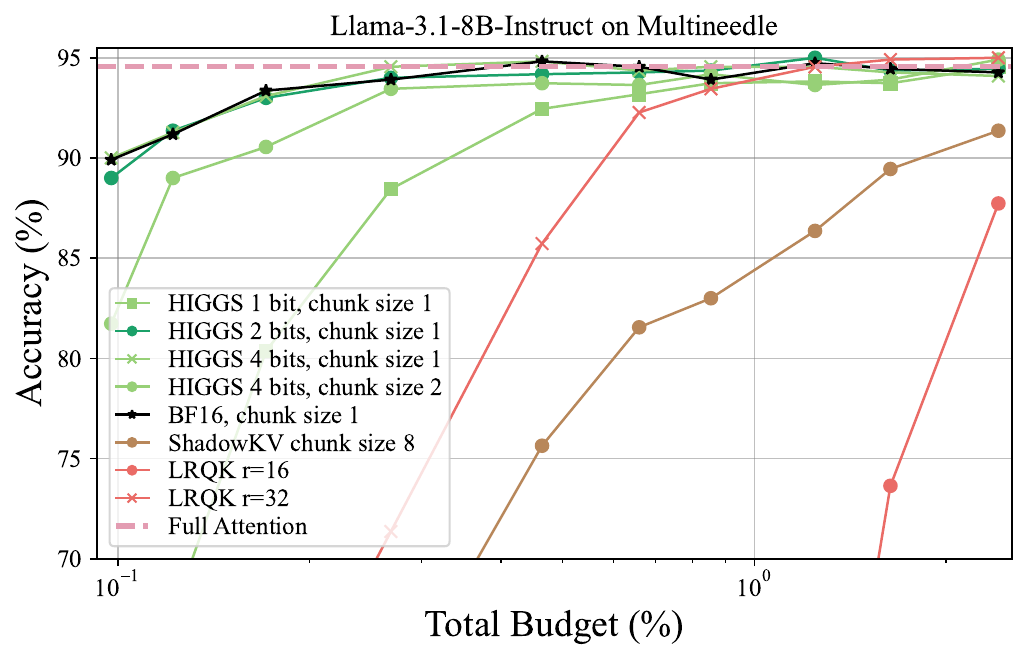}
    \includegraphics[width=0.5\linewidth]{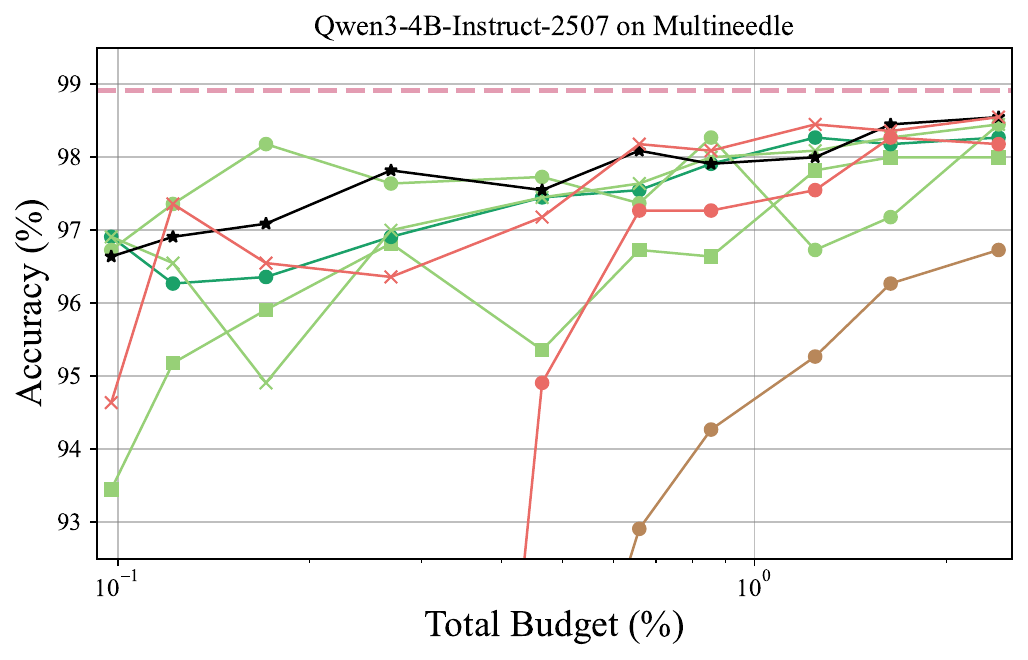}
    \vspace{-10px}
    \caption{Evaluation of ShadowKV offloading (w/o key SVD) with different KV selection types on MultiNeedle-128K for \textbf{(left)} Llama-3.1-8B-Instruct and \textbf{(right)} Qwen3-4B-Instruct-2507. }
    \label{fig:exp_4.3_landmark_quant_multineedle}
    \vspace{-15px}
\end{figure}

\begin{figure}[b]
    \centering
    \vspace{-19px}
    \hspace{-12px}\includegraphics[width=0.5\linewidth]{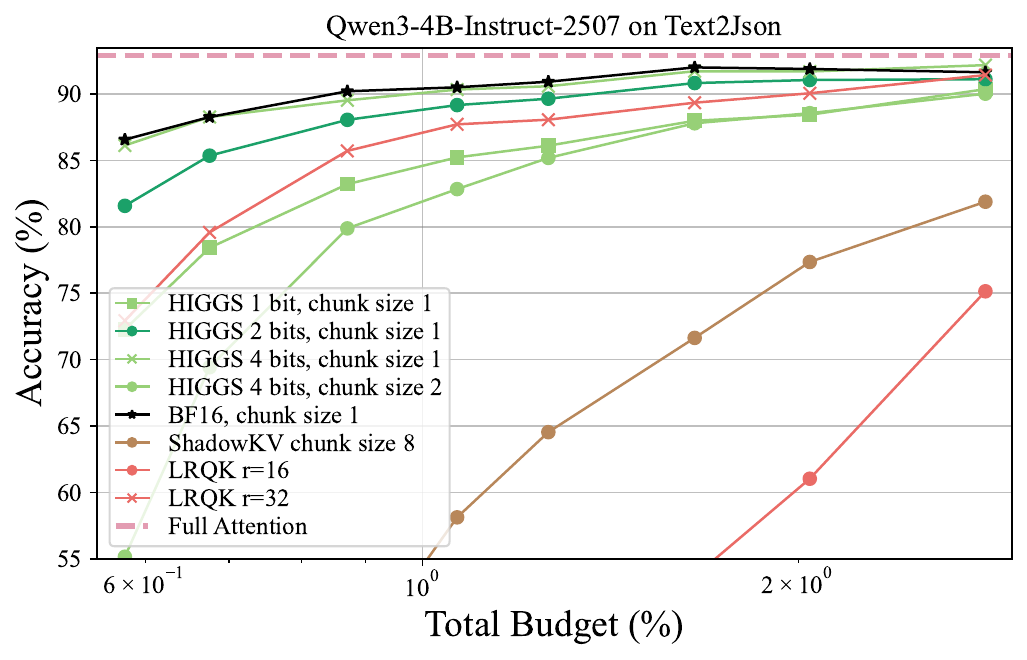}
    \includegraphics[width=0.5\linewidth]{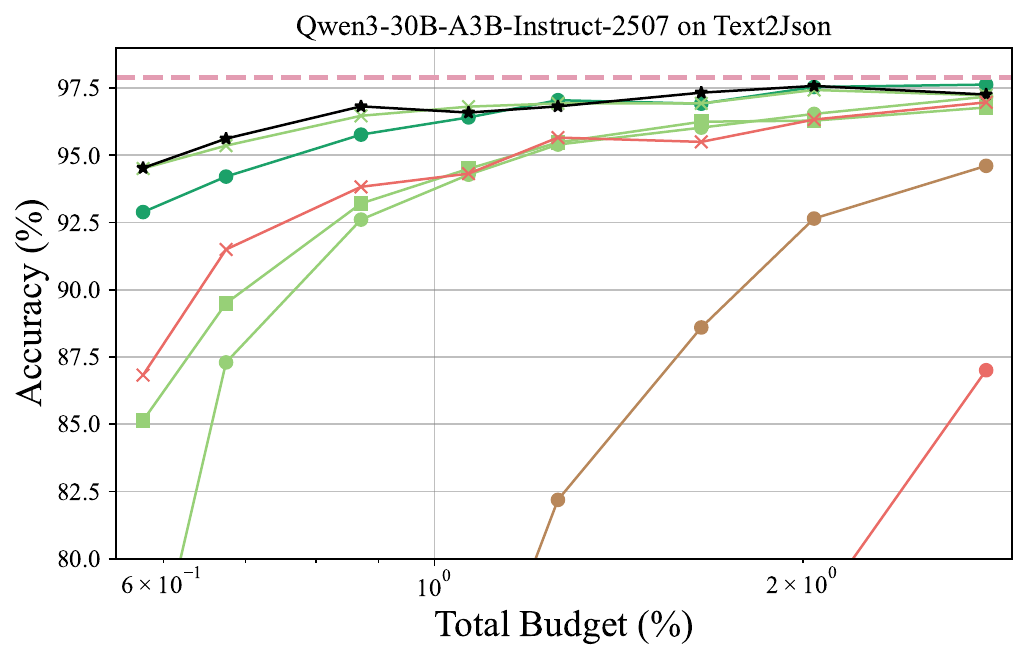}
    \vspace{-10px}
    \caption{Evaluation of ShadowKV offloading (w/o key SVD) with different KV selection type on Text2JSON for \textbf{(left)} Qwen3-4B-Instruct-2507 and \textbf{(right)} Qwen3-30B-A3B-Instruct-2507.}
    \label{fig:exp_4.3_landmark_quant_text2json}
    \vspace{-15px}
\end{figure}

Our previous findings indicate that chunk-based KV selection is inaccurate on context-intensive tasks, even with reduced group size. However, decreasing chunk size to 1 would mean storing all keys on GPU and only offloading values\footnote{Alternatively, keys can be stored in RAM, but then \textit{all} keys must be loaded on each forward pass, which is too slow.}, which would not be practical in real-world offloading setups. This raises the question of how to balance model quality against the amount of data loaded into memory.

Instead of selecting keys as groups, a more straightforward approach is to select individual keys based on compact representations. InfiniGen~\cite{lee2024infinigen} uses compact SVD-based key representations to select which KV entries are loaded from system memory. In subsequent work, LRQK~\cite{tenghui2026efficient} uses a more specialized low-rank approximation to also select which top-k keys are needed at a given step. We also consider a simpler alternative: using extreme quantization. We use the same HIGGS quantizer as in Section~\ref{sect:experiments_svd}, but with 1-2 bit grids ($d{=}8$, $n{=}256$ for 1.02 bit and $d{=}4$,$n{=}256$ for 2.02 bit).


In Figures~\ref{fig:exp_4.3_landmark_quant_multineedle} and~\ref{fig:exp_4.3_landmark_quant_text2json}, we compare ShadowKV group landmarks, LRQK individual low-rank proxies and HIGGS data-free quantization. We compare quantization setups that use the same amount of GPU memory as ShadowKV landmarks. Specifically, the configurations 16-bit/chunk8, 4-bit/chunk2, and 2-bit/chunk1 all require, on average, 2 bits per key on GPU. In turn, BF16/chunk1 serves as an ``oracle'' configuration, providing an upper bound on landmark selection accuracy. Overall, 2-bit HIGGS approximation with chunk size 1 substantially outperforms both ShadowKV (chunk 8, equal GPU memory budget) and LRQK (rank=32) that takes up more GPU memory. In turn, 4-bit HIGGS matches the upper bound performance at similar budgets to LRQK rank 32.

These results suggest that simple data-free quantization allows for more accurate KV selection than both grouping and specialized low-rank representations on context-intensive tasks. Furthermore, not using group landmarks means that KV offloading does not need group outliers (which is why YAKV disables outliers as discussed in Section~\ref{sect:method_yakv_takeaways}), simplifying code and reducing GPU footprint.

\textbf{Further improvements:} we found two more strategies that further improve KV selection:
\begin{enumerate}[leftmargin=*]
    \vspace{-5px}\item \textbf{Residual Landmark Quantization:} we can achieve accurate KV selection at ${\approx}1.5$ bits per value with group-wise RVQ~\cite{jegou2011searching}, i.e. keeping one 4-bit ``landmark'' per 8 keys and 1-bit ``error correction'' for each key in the group. We describe and evaluate this technique in Appendix~\ref{app:residual_quantization}.
    \vspace{-3px}\item \textbf{Adaptive ``Top-P'' budget:} instead of loading a fixed number of KVs per head, we can partition KV budget between heads to maximize coverage, inspired by AdaKV~\cite{feng2024ada,lin2025twilightadaptiveattentionsparsity}, in Appendix~\ref{app:top_p_and_top_kp}.
\end{enumerate}\vspace{-5px}
Note that \textbf{we do not include either in YAKV} to keep our method simple and easy to implement.
\pagebreak

\vspace{-8px}
\subsection{Combined Evaluations}\label{sect:experiments_maintable}
\vspace{-7px}

\begin{table*}[t]
\centering
\vspace{-30px}
\small
\setlength{\tabcolsep}{5pt}
\renewcommand{\arraystretch}{1}
\caption{Evaluation of KV Offloading for Qwen3-Instruct-2507 4B and 30B-A3B on context-intensive tasks, compared in terms of accuracy and average host-to-device transfer per decoding step.}
\label{tab:4.4_qwen3-instruct-2507}
\vspace{-5px}
\begin{tabular}{lcc cc cc cc cc}
\toprule
\multirow{2}{*}{\textbf{Method}}
& \multicolumn{2}{c}{\textbf{Text2JSON}} 
& \multicolumn{2}{c}{\textbf{MultiNeedle}} 
& \multicolumn{2}{c}{\textbf{LongProc}} 
& \multicolumn{2}{c}{\textbf{Loong}} 
& \multicolumn{2}{c}{\textbf{Average}}  \\

\cmidrule(lr){2-3} 
\cmidrule(lr){4-5} 
\cmidrule(lr){6-7} 
\cmidrule(lr){8-9}
\cmidrule(lr){10-11}

& Acc & GiB 
& Acc & GiB 
& Acc & GiB 
& Acc & GiB 
& Acc & GiB \\

\midrule
\multicolumn{11}{c}{Model: \texttt{Qwen/Qwen3-4B-Instruct-2507}} \\
\midrule
\rowcolor{gray!20} Original    & 91.25 & 2.74 & 98.91 & 18.00 & 50.5  & 5.22 & 46.81 & 11.53 & 71.87 & 9.37 \\
ShadowKV    & 69.38 & 0.02 & 92.64 & 0.14  & 15.75 & 0.04  & 44.56 & 0.09 & 55.58 & 0.07 \\
ArkVale     & 22.10 & 0.02 & 96.64 & 0.14  & 3.37     & 0.04     & -     & 0.09 & - & - \\
LRQK        & 87.99 & 0.02 & 98.64 & 0.14  & 8.75  & 0.04  & 44.70 & 0.09 & 60.02 & 0.07 \\
YAKV (ours) & 90.83 & 0.02 & 98.45 & 0.14  & 33.37 & 0.04  & 47.34 & 0.09 & \textbf{67.50} & 0.07 \\

\midrule
\multicolumn{11}{c}{Model: \texttt{Qwen/Qwen3-30B-A3B-Instruct-2507}} \\
\midrule
\rowcolor{gray!20} Original    & 97.88 & 1.84 & 100.00& 12.00 & 59.7  & 3.48 & 56.29 & 7.69 & 78.47 & 6.25 \\
ShadowKV    & 80.39 & 0.014 & 98.82 & 0.09  & 27.00 & 0.03  & 51.53 & 0.06 & 64.44 & 0.05 \\
ArkVale     & 46.05 & 0.014 & 99.91 & 0.09  & 14.06 & 0.03  & -     & 0.06 & - & - \\
LRQK        & 95.66 & 0.014 & 99.82 & 0.09  & 31.75     & 0.03     & 55.50 & 0.06 & 70.68 & 0.05 \\
YAKV (ours) & 96.74 & 0.014 & 99.82 & 0.09  & 38.00 & 0.03  & 57.58 & 0.06 & \textbf{73.04} & 0.05 \\

\bottomrule
\end{tabular}
\vspace{-15px}
\end{table*}

\begin{table*}[b]
\centering
\vspace{-15px}
\small
\setlength{\tabcolsep}{5pt}
\renewcommand{\arraystretch}{1}
\caption{Evaluation of KV Offloading for Llama 3.1-8B and 3.2-3B (both Instruct) on context-intensive tasks, compared in terms of accuracy and average host-to-device transfer per decoding step.}
\label{tab:4.4_llama-3.x-instruct}
\vspace{-5px}
\begin{tabular}{lcc cc cc cc cc}
\toprule
\multirow{2}{*}{\textbf{Method}}
& \multicolumn{2}{c}{\textbf{Text2JSON}} 
& \multicolumn{2}{c}{\textbf{MultiNeedle}} 
& \multicolumn{2}{c}{\textbf{LongProc}} 
& \multicolumn{2}{c}{\textbf{Loong}} 
& \multicolumn{2}{c}{\textbf{Average}}  \\

\cmidrule(lr){2-3} 
\cmidrule(lr){4-5} 
\cmidrule(lr){6-7} 
\cmidrule(lr){8-9}
\cmidrule(lr){10-11}

& Acc & GiB 
& Acc & GiB 
& Acc & GiB 
& Acc & GiB 
& Acc & GiB \\

\midrule
\multicolumn{11}{c}{Model: \texttt{meta-llama/Llama-3.1-8B-Instruct}} \\
\midrule

\rowcolor{gray!20} Original    & 46.75 & 2.44 & 94.55 & 16.00 & 22.71  & 4.64 & 37.28 & 9.64 & 50.32 & 8.18 \\
ShadowKV    & 10.46 & 0.02 & 85.73 & 0.12  & 4.59  & 0.04  & 31.25 & 0.08 & 33.01 & 0.06 \\
InfiniGen   & 2.64  & 0.02 & 84.91 & 0.12  & 0.00  & 0.04  & - & 0.08 & - & 0.06 \\
ArkVale     & 1.19 & 0.02 & 90.09 & 0.12  & 0.06  & 0.04  & - & 0.08 & - & 0.06 \\
LRQK        & 3.30  & 0.02 & 93.18 & 0.12  & 0.85  & 0.04  & 30.49  & 0.08 & 31.96 & 0.06 \\
YAKV (ours) & 30.60 & 0.02 & 93.36 & 0.12  & 14.02 & 0.04  & 33.53 & 0.08 & \textbf{42.88} & 0.06 \\

\midrule
\multicolumn{11}{c}{Model: \texttt{meta-llama/Llama-3.2-3B-Instruct}} \\
\midrule

\rowcolor{gray!20} Original    & 27.69 & 2.14 & 85.45 & 14.00 & 0.99  & 4.06 & 20.77 & 8.44 & 33.73 & 7.16 \\
ShadowKV    & 9.46  & 0.02 & 72.91 & 0.11  & 0.15  & 0.03  & 19.74 & 0.07 & 25.57 & 0.06 \\
InfiniGen   & 5.47  & 0.02 & 84.73 & 0.11  & 0.01  & 0.03  & - & 0.07 & - & 0.06 \\
ArkVale     & 0.45  & 0.02 & 78.82 & 0.11  & 0.01  & 0.03     & - & 0.07 & - & 0.06 \\
LRQK        & 2.05  & 0.02 & 84.64 & 0.11  & 0.01  & 0.03  & 18.31 & 0.07 & 26.25 & 0.06 \\
YAKV (ours) & 25.70 & 0.02 & 80.82 & 0.11  & 0.24  & 0.03  & 19.84 & 0.07 & \textbf{31.65} & 0.06 \\
\bottomrule
\end{tabular}

\vspace{-10px}
\end{table*}

In this section, we evaluate our previous observations across a broader range of models and context-intensive tasks. We use the four benchmarks identified earlier: Text2JSON, MultiNeedle-128K~\cite{li2025needlebench}, LongProc HTML to TSV 8K~\cite{ye25longproc}, and Loong~\cite{wang2024leave}. We evaluate four LLMs from two popular families: Qwen3-4B-Instruct-2507, Qwen3-30B-A3B-Instruct-2507~\cite{yang2025qwen3technicalreport}, Llama 3.1 8B, and Llama 3.2 3B~\cite{touvron2023llama,dubey2024llama}.
We compare the following KV offloading methods:\begin{itemize}[leftmargin=*]
    \vspace{-5px}\item \textbf{YAKV}~(ours): offloading with 4-bit HIGGS compression and 2-bit HIGGS for selection.
    \vspace{-3px}\item \textbf{ShadowKV}~\cite{sun2025shadowkv}: offloading with SVD compression, group channel-mean landmarks and outliers.
    \vspace{-3px}\item \textbf{ArkVale}~\cite{chen2024arkvale}: offloading with group-based KV selection using cuboid-mean approximations.
    \vspace{-3px}\item \textbf{LRQK}~\cite{tenghui2026efficient}: offloading with specialized low-rank decomposition for uncompressed key selection.
    \vspace{-3px}\item \textbf{InfiniGen}~\cite{lee2024infinigen}: offloading with individual SVD-based KV selection. Does not support Qwen3.
\end{itemize}\vspace{-5px}

For fair comparison, we evaluate these methods using the same total PCIe budget per token (GiB). More specifically, we use the recommended ShadowKV parameters as our reference point and tune YAKV, LRQK and InfiniGen to fit into the budget, using recommended parameters where possible. One notable exception is that InfiniGen does not natively support Group Query Attention (GQA~\cite{ainslie2023gqa}). To circumvent this, we modified the official implementation to aggregate key scores within the same query group, similar to ShadowKV. We provide detailed configurations in Appendix~\ref{app:baseline_configs_and_extras}.

The results in Tables~\ref{tab:4.4_qwen3-instruct-2507}~\&~\ref{tab:4.4_llama-3.x-instruct} align with our earlier findings with a few caveats. For Qwen3 models, YAKV consistently achieves near-lossless accuracy and outperforms baselines in most cases. LRQK scores slightly (0.09\%) higher on MultiNeedle, but drops significantly on other benchmarks, with other baselines faring worse. Llama 3 shows similar results, except that all methods score poorly on Text2JSON and LongProc. Upon closer inspection, we found that this is caused by the weaker model getting ``stuck'' and generating invalid output structures (see additional setups in Appendix~\ref{app:yakv_extra_evals}).

\pagebreak
\vspace{-5px}
\subsection{GPU Inference Throughput}\label{sect:experiments_inference}
\vspace{-5px}

\begin{table}[t]
\centering
\setlength{\tabcolsep}{3pt}
\caption{Inference throughput evaluation using a single H200 GPU for \textbf{(left)} synthetic 65K prompts with forced decoding and \textbf{(right)} real data, including prefill time with continuous batching.}
\label{tab:4.5_inference}
\vspace{2px}
\small

\begin{minipage}[t]{0.45\linewidth}
\centering
\resizebox{1.05\linewidth}{!}{
\begin{tabular}{lccccc}
\toprule
\textbf{Setup} & \textbf{Batch} & \textbf{TPOT (ms)} & \textbf{Throughput (tok/s)} & \textbf{Rel. Speedup} \\
\midrule
\multicolumn{5}{c}{\texttt{Qwen3-30B-A3B-Instruct-2507}, synthetic data} \\
\midrule
Baseline & 1 & 54ms & 18.5 & 1.00 \\
YAKV & 16 & 380ms & \textbf{42.1} & \textbf{2.28} \\
\midrule
\multicolumn{5}{c}{\texttt{Qwen3-32B (YARN)}, synthetic data} \\
\midrule
Baseline & 1 & 37ms & 27.0 & 1.00 \\
YAKV & 16 & 370ms & \textbf{43.2} & \textbf{1.60} \\
\bottomrule
\end{tabular}
}
\end{minipage}
\hfill
\begin{minipage}[t]{0.45\linewidth}
\centering
\resizebox{1.05\linewidth}{!}{
\begin{tabular}{lccccc}
\toprule
\textbf{Setup} & \textbf{Batch} & \textbf{TPOT (ms)} & \textbf{Throughput (tok/s)} & \textbf{Rel. Speedup} \\
\midrule
\multicolumn{5}{c}{\texttt{Qwen3-30B-A3B-Instruct-2507}, real data} \\
\midrule
Baseline & 1 & 38ms & 26.32 & 1.00 \\
YAKV & 32 & 420ms & \textbf{76.2} & \textbf{2.89} \\
\midrule
\multicolumn{5}{c}{\texttt{Qwen3-30B-A3B-Thinking-2507}, real data} \\
\midrule
Baseline & 1 & 39ms & 25.64 & 1.00 \\
YAKV & 32 & 252ms & \textbf{127.0} & \textbf{4.95} \\
\bottomrule
\end{tabular}
}
\end{minipage}
\hspace{15px}
\vspace{-15px}
\end{table}

In this section, we evaluate real-world GPU inference throughput to verify that our KV selection techniques are practical. We implement a minimal version of YAKV offloading on top of mini SGLang~\cite{zheng2024efficiently} codebase with modified HIGGS kernels~\cite{flute2024,malinovskii2024pushing} (see Appendix~\ref{app:inference_experiments} for details) and evaluate inference throughput in two setups: synthetic 65K prompts (forced decoding w/o EOS) and shuffled Text2JSON+MultiNeedle inputs (realistic decoding). For a realistic evaluation scenario, we continuously add new client requests as the model finishes processing existing ones. We use SGLang continuous batching: the new prompts are  encoded and added to the current batch on the fly.

We report the optimal batch size that fits in GPU memory, individual token generation latency (TPOT, time per output token), and the overall inference throughput (tokens per second). Both Throughput and TPOT include the prefill time which is unaffected by offloading. For YAKV, we use the same budget parameters as in Section~\ref{sect:experiments_maintable}. For synthetic inputs, we evaluate Qwen3-30B-A3B-Instruct-2507 (MoE) and dense Qwen3-32B  models. The 32B model natively supports sequence length up to 32,768, so we use the recommended YARN~\cite{yarn} configuration to support longer sequences\footnote{Context size and YARN from model card, \url{https://huggingface.co/Qwen/Qwen3-32B}.}. For real data, we compare Qwen3-30B-A3B Instruct and Thinking variants: these have the same number of parameters, but differ in how many \textit{output} tokens they generate per prompt. The Thinking model generates more tokens, making it a decoding-heavy scenario, while the Instruct version generates less and spends a larger portion of time on prefill (increasing average TPOT).

The results in Table~\ref{tab:4.5_inference} demonstrate that the offloaded inference can achieve $1.5{-}4.9{\times}$ higher throughput than full attention by processing larger input batches that would otherwise not fit into memory. This is consistent with previous works on KV offloading, showing that YAKV can maintain high inference throughput while being more accurate on context-intensive tasks (see Section~\ref{sect:experiments_maintable}). We provide more detailed configurations and evaluation results (e.g. TTFT) in Appendix~\ref{app:inference_experiments}.

\vspace{-10px}
\section{Discussion}\label{sect:discussion}
\vspace{-8px}

In this work, we tested the limits of sparse KV offloading and found that simply increasing the amount of required information makes the problem harder for offloading.
We gathered a dataset of context-intensive tasks and combine it with context-intensive subsets of existing benchmarks to test modern offloading methods.
Our findings demonstrate that modern KV offloading algorithms become lossy or require significantly higher PCIe budgets.

However, unlike traditional KV eviction, \textit{offloading is fundamentally capable of solving context-intensive tasks.} The failures we observed can be attributed to 1) overly aggressive key compression that worked on easier problems and 2) inaccurate token selection heuristics based on landmarks.
These are not conceptual problems with the offloading itself, but technical limitations that can be circumvented with better compression and key selection. Furthermore, our YAKV evaluations demonstrate that addressing these two failure modes can significantly improve KV offloading accuracy while retaining high inference throughput. We hope that our findings and evaluation sets will help inform the design of future KV offloading methods
\footnote{Data and code available at \url{https://github.com/yandex-research/context-intensive-kv-offloading}}.

\textbf{Limitations.} The main focus of our work is to better understand the effectiveness of KV offloading, not design the highest performance algorithm. As such, we designed YAKV as a minimal working KV offloading system without complex technical features such as KV prefetch, adaptive budgeting or better quantization that could improve performance (see Appendix~\ref{app:residual_quantization}~\&~\ref{app:top_p_and_top_kp}).
Our analysis is also limited to models trained with full attention: models with alternative attention types such as MLA~\cite{deepseek_mla_2024} or Gated DeltaNet~\cite{yang2024gated} present interesting directions for future research.

\textbf{Acknowledgements.} Authors thank Vladislav Kruglikov for brainstorming about practical LLM inference concerns and GPU implementation matters. We also thank Irina Lialikova from Yandex LLM Analytics team for helpful discussions about production workloads. Finally, we thank Gleb Rodionov for his suggestions about how to analyze accuracy drawdown in different models.

\bibliography{main}
\bibliographystyle{unsrt}

\pagebreak
\appendix
\vspace{-8px}
\section{Preliminary Benchmark Exploration \& Configurations}\label{app:preliminary_experiments}
\vspace{-7px}
Before our primary investigation in Sections~\ref{sect:method}~\&~\ref{sect:experiments}, we ran preliminary experiments on a range of popular benchmarks with KV offloading. Notably, the main results of ShadowKV reproduce easily and perfectly from the official code: Llama 3.1 8B consistently scores near-losslessly on RULER and select LongBench subtasks. We then apply the same code to evaluate on more context-intensive benchmarks: Loong~\cite{wang2024leave} and MultiNeedle. We use the official Loong codebase\footnote{\url{https://github.com/mozerwang/loong}} using \texttt{gpt-4-turbo} for LLM-as-a-judge. The results for Loong are summarized in Table~\ref{tab:app_loong}: the context-intensive benchmark (``Leave No Document Behind'') shows significant accuracy drawdowns across all levels with default ShadowKV hyperparameters. This aligns with our observations and makes Loong an important example of context-intensive multi-document QA. The only reason why we prefer MultiNeedle for Sections~\ref{sect:experiments_svd}---\ref{sect:experiments_better_landmarks} is that evaluating on Loong incurs significant API costs for LLM-as-a-Judge, making it prohibitive to run extensive budget sweeps necessary for our analysis.

\begin{table}[h]
\vspace{-15px}
\centering
\caption{Comparison of Full Attention and ShadowKV on Loong for Llama-3.1-8B-Instruct. Level 1: Spotlight Locating, level 2: Comparison, Level 3: Clustering, Level 4: Chain of Reasoning}
\label{tab:app_loong}
\vspace{3px}
\begin{tabular}{lcccccccc}
\toprule
 & \multicolumn{2}{c}{Level 1} 
 & \multicolumn{2}{c}{Level 2} 
 & \multicolumn{2}{c}{Level 3} 
 & \multicolumn{2}{c}{Level 4} \\
\cmidrule(lr){2-3}
\cmidrule(lr){4-5}
\cmidrule(lr){6-7}
\cmidrule(lr){8-9}
Method 
 & Score & Perfect 
 & Score & Perfect 
 & Score & Perfect 
 & Score & Perfect \\
\midrule
Full Attention 
 & 63.24 & 0.53 
 & 37.20 & 0.19 
 & 25.63 & 0.01 
 & 37.27 & 0.18 \\
ShadowKV 
 & 46.15 & 0.37 
 & 25.81 & 0.12 
 & 15.70 & 0.01 
 & 24.75 & 0.06 \\
\bottomrule
\end{tabular}
\vspace{-10px}
\end{table}

Our MultiNeedle evaluations use the MultiNeedle Retrieval 128K configuration from NeedleBench v2~\cite{li2025needlebench} using OpenCompass (commit id \href{https://github.com/open-compass/opencompass/commit/12462107fe746db536bc4b44bb6b58f0736251fe}{\texttt{1246210}}). This setup hides 11 synthetic ``needles'' among 128K prompt tokens and scores how many of those the model finds correctly (exact match accuracy). We use \href{https://github.com/open-compass/opencompass/blob/12462107fe746db536bc4b44bb6b58f0736251fe/opencompass/configs/datasets/needlebench_v2/needlebench_v2_128k/needlebench_v2_multi_retrieval_128k.py}{\texttt{needlebench\_v2\_multi\_retrieval\_128k}} configuration. For experimental consistency, we use the same sample of 100 english samples instead of drawing random needles for each experiment.

\begin{table*}[h!]
\centering
\vspace{-10px}
\caption{Generation and extraction prompts used for each data type in Text2JSON.}
\label{tab:text2json-prompts}
\vspace{-5px}
\small
\setlength{\tabcolsep}{4pt}
\renewcommand{\arraystretch}{1.2}
\begin{tabularx}{\textwidth}{
    >{\raggedright\arraybackslash}p{1.5cm}
    >{\raggedright\arraybackslash}X
    >{\raggedright\arraybackslash}X
    >{\raggedright\arraybackslash}X
    >{\raggedright\arraybackslash}X
}
\toprule
\textbf{Subset} & \textbf{Doctors} & \textbf{Movies} & \textbf{Organizations} & \textbf{Products} \\
\midrule

\textbf{Generation} &
Generate $X$ entries, each starting with a newline, in the following format: \newline
\texttt{Name Surname, Doctor Specialization, City}
&
Suggest $X$ unique movies, each starting with a newline, in the following format: \newline
\texttt{Movie title, Country of Production, Year of Production}
&
Generate 200 organization cards, each starting with a newline, in the following format: \newline
\texttt{The name of the organization, address, site}
&
Generate 10 product cards in the following format: \newline
\texttt{Product name: <Product name>} \newline
\texttt{* Color: <Color>} \newline
\texttt{* Material: <Material>} \newline
\texttt{* Length: <Length>} \newline
\texttt{* Category: <Category>}
\\

\midrule

\textbf{Extraction} &
Find all doctor review cards in the text and compose a JSON object with the following fields: \texttt{name} --- doctor's name; \texttt{specialization} --- specialization; \texttt{city} --- city. There is no need to reproduce the reviews. Output only JSON. Do not skip cards and do not produce duplicates.
&
Find all movie review cards in the text and compose a JSON object with the following fields: \texttt{name} --- movie title; \texttt{country} --- country of production; \texttt{year} --- year of release. There is no need to reproduce the reviews. Output only JSON. Do not skip cards and do not produce duplicates.
&
Find all organization cards in the text and compose a JSON object with the following fields: \texttt{name} --- the name of the organization (exactly as written in the card); \texttt{address} --- the address; \texttt{site} --- the website. There is no need to reproduce the reviews. Output only JSON. Do not skip cards and do not produce duplicates.
&
Find all product cards in the text and compose a JSON object with the following fields: \texttt{name} --- product name (exactly as written in the card); \texttt{material} --- material; \texttt{color} --- color. There is no need to reproduce the descriptions. Output only JSON. Do not skip cards and do not produce duplicates.
\\

\bottomrule
\end{tabularx}
\vspace{-27.5px}
\end{table*}

\pagebreak

\vspace{-5px}
\section{Text2JSON construction and evaluation}\label{app:text2json_construction}
\vspace{-5px}

Text2JSON is constructed from four types of entries: organization cards, doctor cards, movie review cards, and product cards. All entries are generated by GPT-5.2 using predefined prompts.
Importantly, the benchmark contains no personal data. To build each benchmark instance, we randomly sample between 3 and 20 entries from each category and between 3 and 10 passages from FineWeb-Edu \cite{penedo2024finewebdatasetsdecantingweb} and moderate the results manually. These segments are then concatenated using \texttt{\textbackslash n\textbackslash n} as a separator. We provide specific prompts for each type in Table~\ref{tab:text2json-prompts}.

Given such an input, the model is required to produce a valid JSON object containing all entries of the target type, in arbitrary order.
We avoid LLM-as-a-Judge and use a deterministic, name-anchored metric.  
Each predicted/gold record is a dictionary with a unique name key and two additional fields.  
Evaluation first aligns prediction and gold by exact name match. Unmatched predicted names are treated as false positives, and unmatched gold names are false negatives.  
Each matched entry receives a score in \(0,1\): it gets 1.0 when all required fields are present and all values are correct, and it is reduced when fields are missing or values are incorrect. These per-entry scores are summed and then normalized by the total number of matched entries plus false positives and false negatives.  
This is a soft IoU-style entity metric: the denominator penalizes missing/spurious entities (set overlap behavior), while the numerator gives partial credit for attribute correctness on matched entities.

\section{Additional Experiments for Section~\ref{sect:experiments_svd}}\label{app:experiments_svd}
\begin{figure}[h]
    \centering
    \vspace{-15px}
    \hspace{-12px}\includegraphics[width=0.51\linewidth]{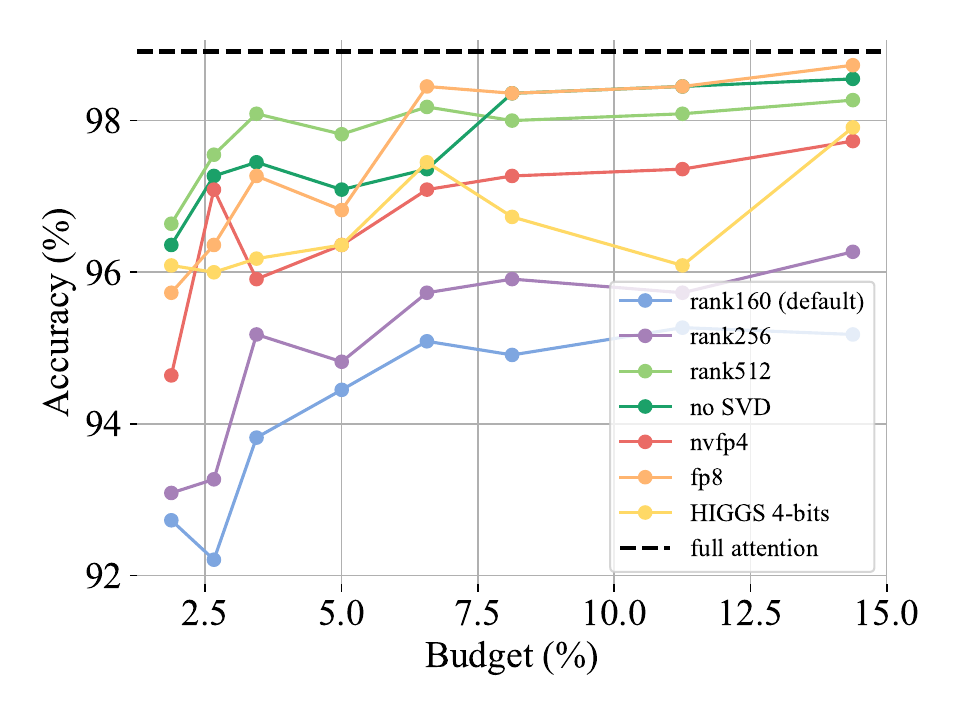}
    \includegraphics[width=0.51\linewidth]{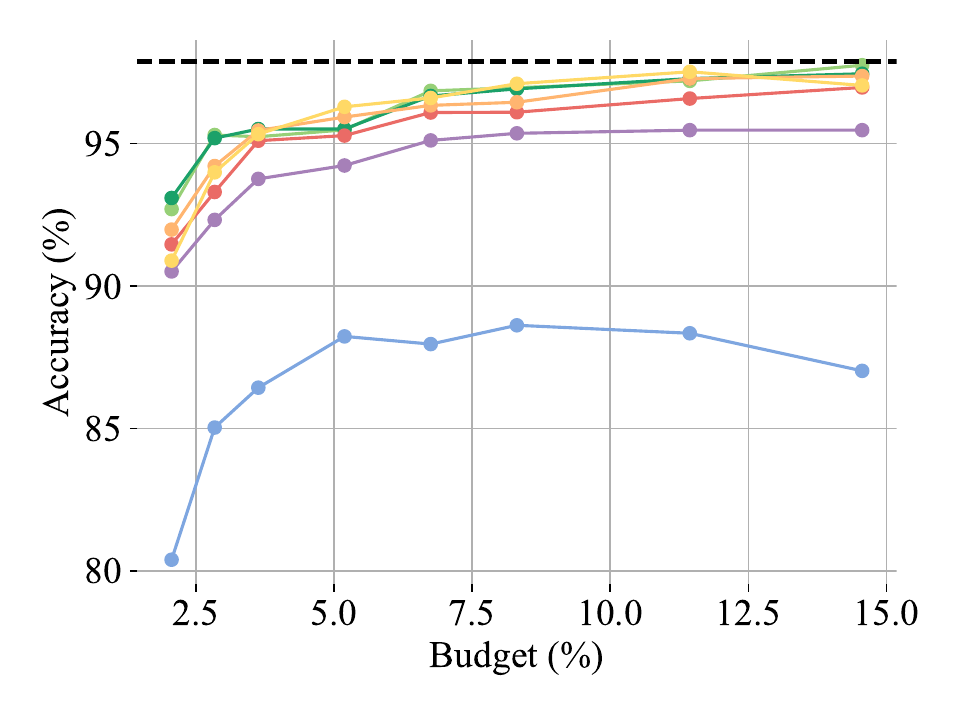}\\[-12pt]
    \hspace{-10px}
    \includegraphics[width=0.51\linewidth]{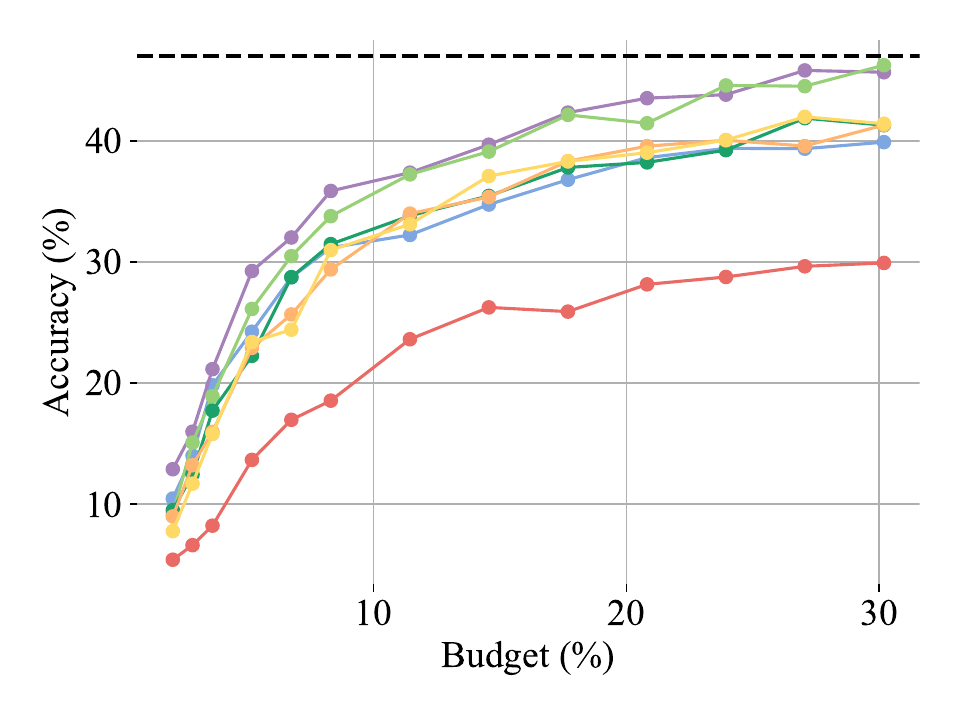}
    
    \vspace{-10px}
    \caption{Evaluation of ShadowKV offloading with different KV compression strategies for Qwen3-4B-Instruct-2507 on MultiNeedle (left), Qwen3-30B-A3B-Instruct-2507 on Text2JSON (right) and Llama3.1-8B on Text2JSON (bottom). The X axis denotes the total percentage of tokens loaded (sparse, outlier and local tokens), see Section~\ref{sect:experiments_svd}.}
    \label{fig:exp_4.1_appendix_svd_quant}
    \vspace{-15px}
\end{figure}

\pagebreak
\section{Additional Experiments for Section~\ref{sect:experiments_budgets_groups}}\label{app:experiments_budgets_groups}

\begin{figure}[h]
    \centering
    \vspace{-15px}
    \hspace{-12px}
    \includegraphics[width=0.51\linewidth]{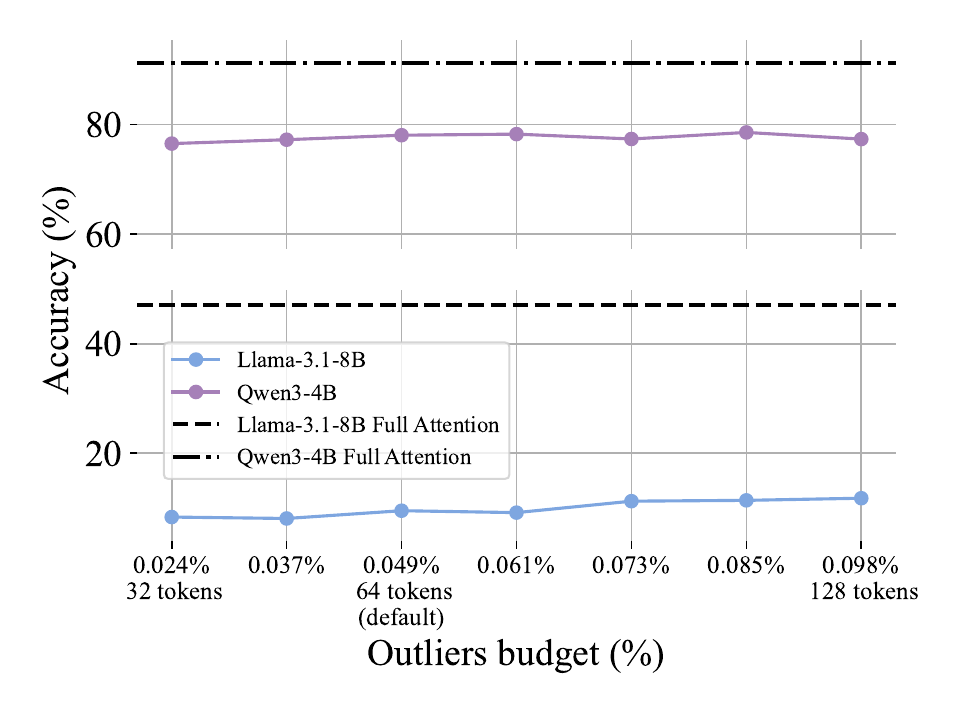}
    \includegraphics[width=0.51\linewidth]{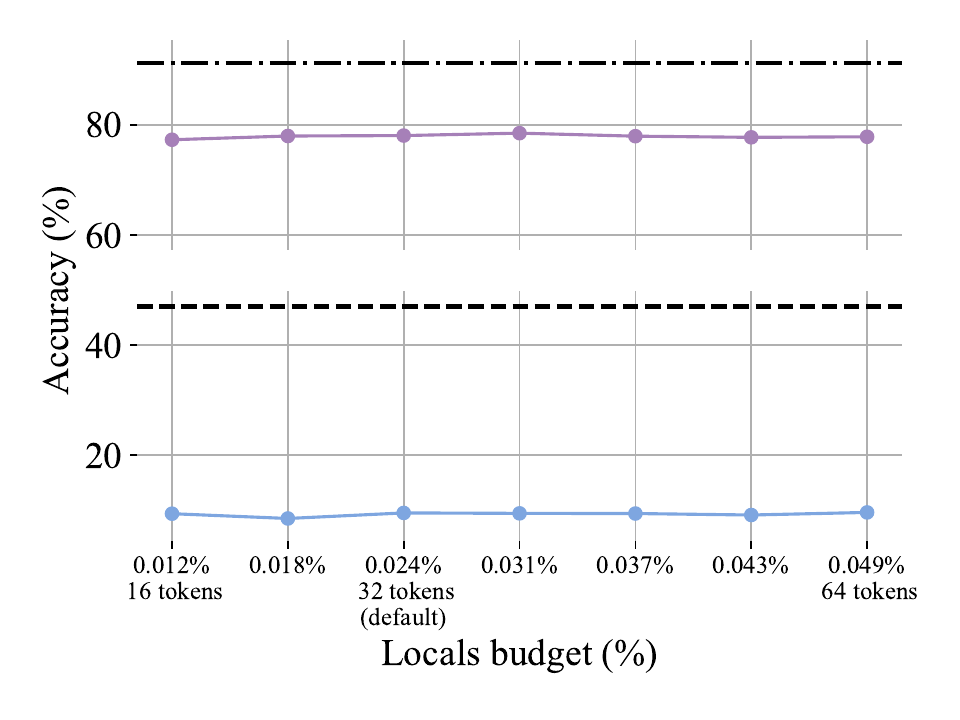}
    \hspace{-10px}
    
    \vspace{-10px}
    \caption{Evaluation of ShadowKV (w/o SVD compression) offloading for Section~\ref{sect:experiments_budgets_groups} with varying outlier budget (left) and local window (right) for Llama-3.1-8B-Instruct and Qwen3-4B-Instruct-2507 on Text2Json dataset.}
    \label{fig:exp_4.2_appendix_local_sparse_budgets}
    \vspace{-15px}
\end{figure}

\section{Residual Quantization of Landmarks}\label{app:residual_quantization}

\begin{figure}[h]
    \centering
    \includegraphics[width=0.32\linewidth]{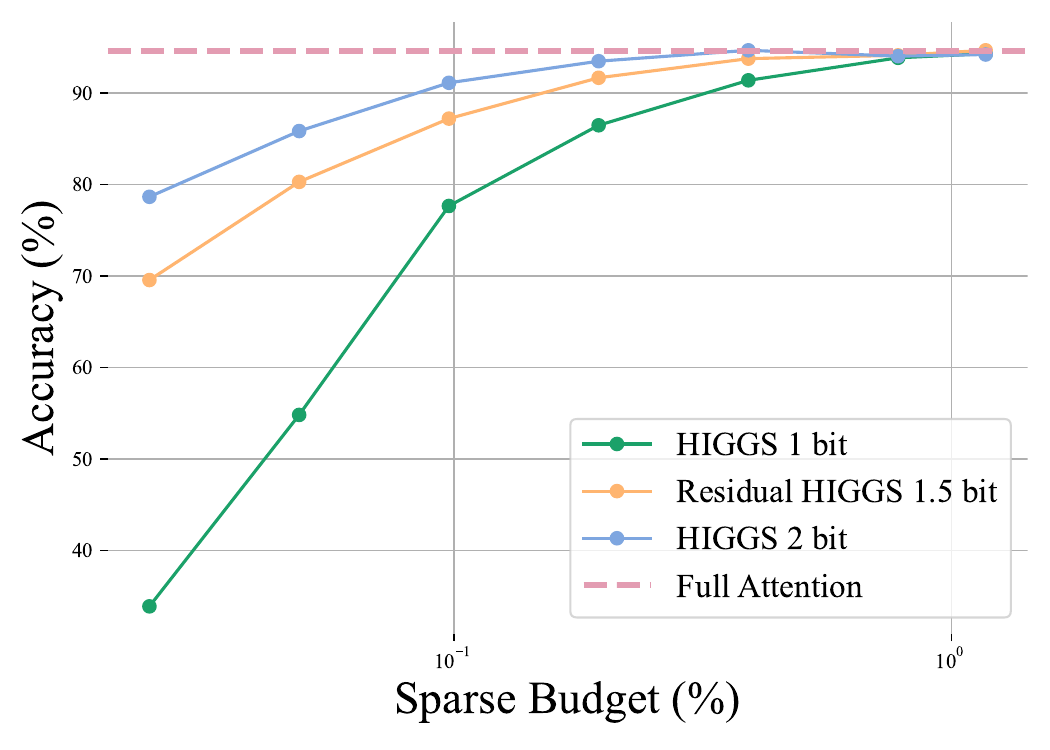}
    \includegraphics[width=0.32\linewidth]{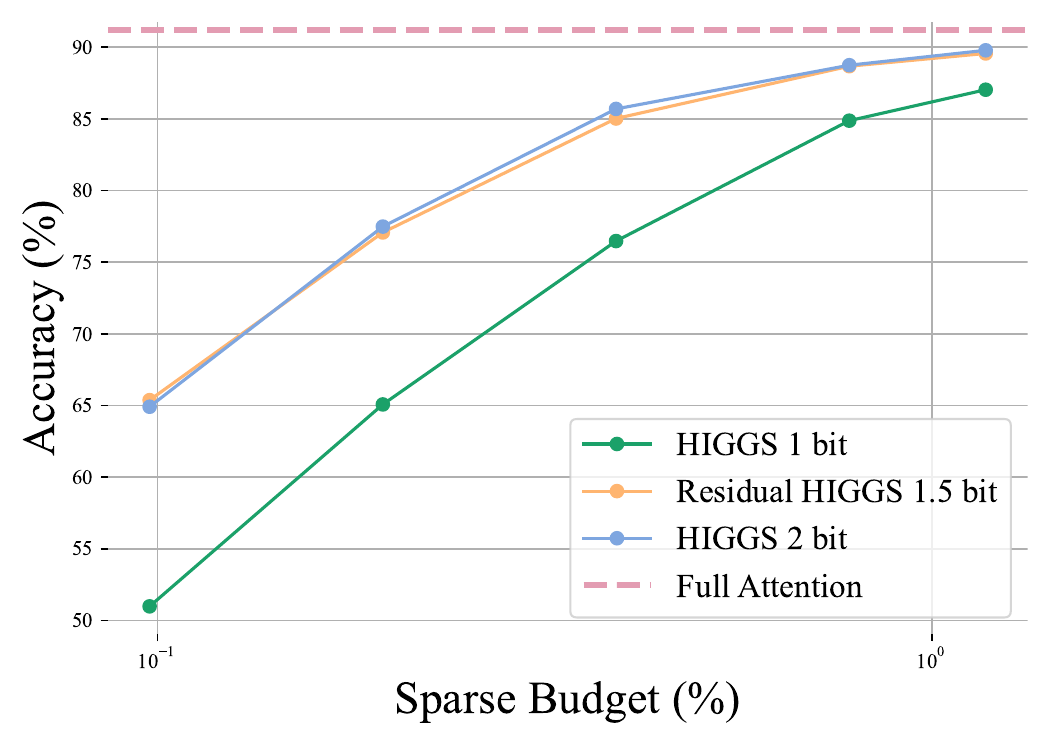}
    \includegraphics[width=0.32\linewidth]{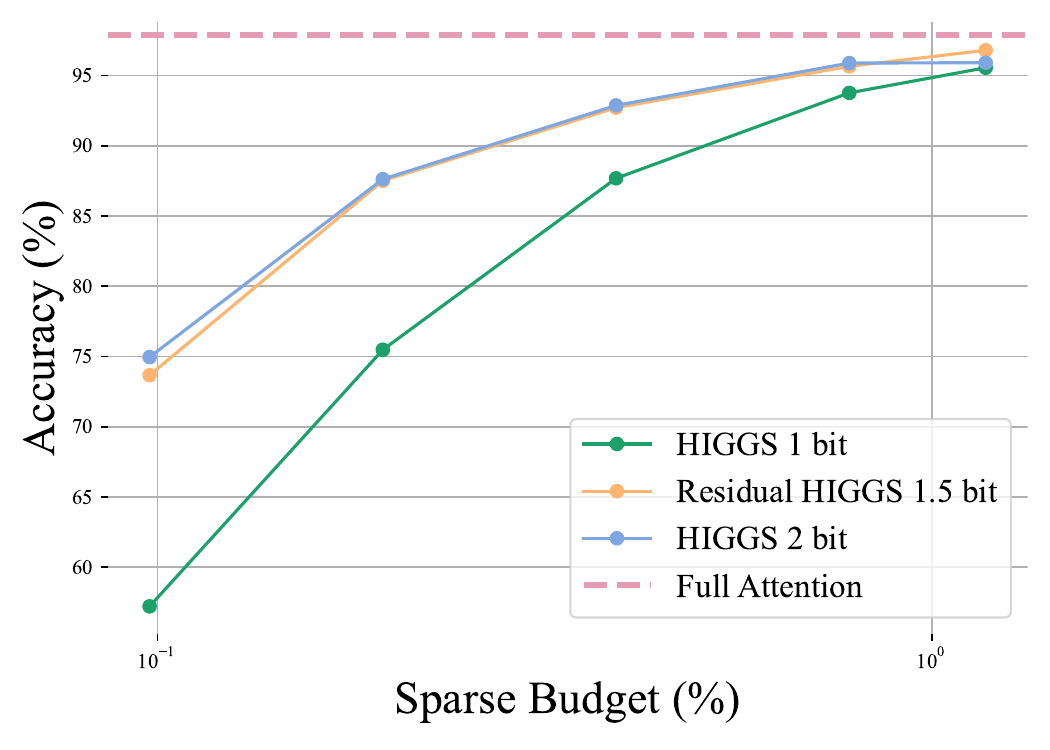}
    \caption{Comparison of 1.5-bit residual landmark quantization with 1-bit and 2-bit HIGGS. Results are shown for Llama-3.1-8B-Instruct on MultiNeedle (left), and for Qwen-3-4B-Instruct-2507 (middle) and Qwen3-30B-A3B-Instruct-2507 (right) on Text2JSON.}
    \label{fig:appendix_residual_quantization}
    \vspace{-15px}
\end{figure}

Low-bit landmark compression can noticeably degrade performance, as shown in Figures~\ref{fig:exp_4.3_landmark_quant_multineedle} and~\ref{fig:exp_4.3_landmark_quant_text2json} for the 1-bit setting. To achieve a better trade-off between the substantial performance degradation of 1-bit quantization and the near-lossless behavior of higher-precision representations, we propose \emph{residual quantization} for landmarks. Specifically, we first construct landmarks at higher precision (e.g., 4-bit HIGGS) using a larger chunk size (the default chunk size of 8 used in ShadowKV), and then quantize the residual between the original keys and the quantized landmarks to a low bit width (1 bit). This configuration has the same memory footprint as 1.5-bit quantized landmarks with a chunk size of 1. We report the evaluation results of residual-quantized landmarks and vanilla HIGGS in Figure~\ref{fig:appendix_residual_quantization}. The results show that residual quantization achieves performance close to that of 2-bit landmarks while incurring 25\% less memory overhead.

Note that residual quantization does not require reconstructing the landmarks in order to compute the highest-scoring dot products. Let $\hat{K} = \texttt{repeat}(L) + R$, where $L$ denotes the chunk-average landmarks and $R$ the low-bit residuals. Then the query--key dot products can be written as
$$
Q \hat{K} = Q(\texttt{repeat}(L) + R) = \texttt{repeat}(QL) + QR.
$$
That is, the dot products can be computed by first multiplying the query by the quantized landmarks, then repeating the resulting scores for all vectors within each chunk, and finally adding the dot product between the query and the residuals. For HIGGS-quantized landmarks, this procedure can be implemented using existing inference kernels~\cite{malinovskii2024pushing}.

An even faster \textit{approximate} top-$k$ procedure can be derived from residual quantization methods used in nearest-neighbor search~\cite{jegou2011searching}, although related ideas had appeared earlier in other forms~\cite{arbabenko_parallel}. In our setting, the algorithm first selects a subset of landmarks with high dot products for the current query and then computes residual contributions only for keys belonging to the chunks associated with those selected landmarks.

\section{Adaptive Budgets with Top-P and Top-KP Loading}\label{app:top_p_and_top_kp}

\begin{figure}[h]
    \centering
    \includegraphics[width=0.49\linewidth]{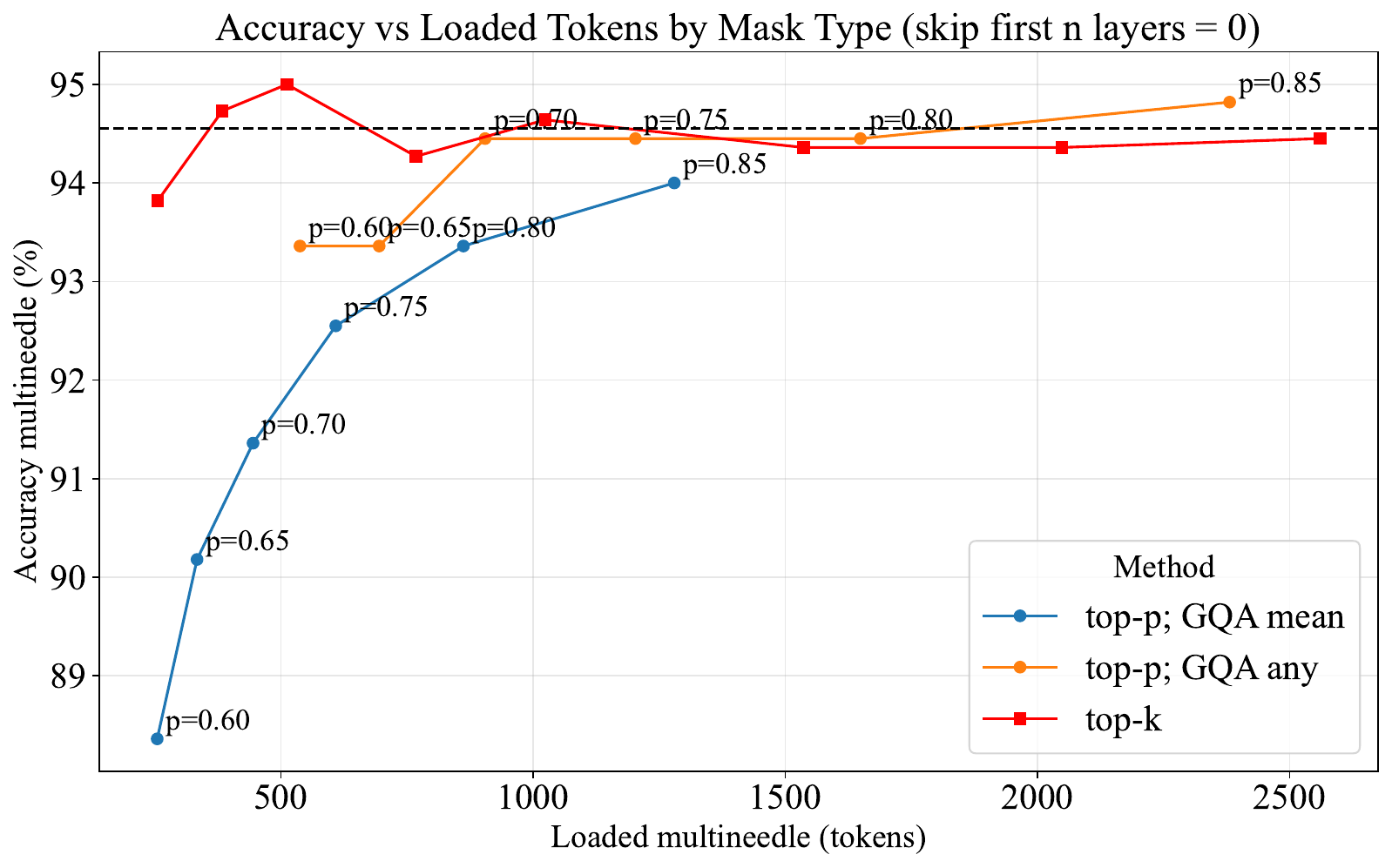}
    \includegraphics[width=0.49\linewidth]{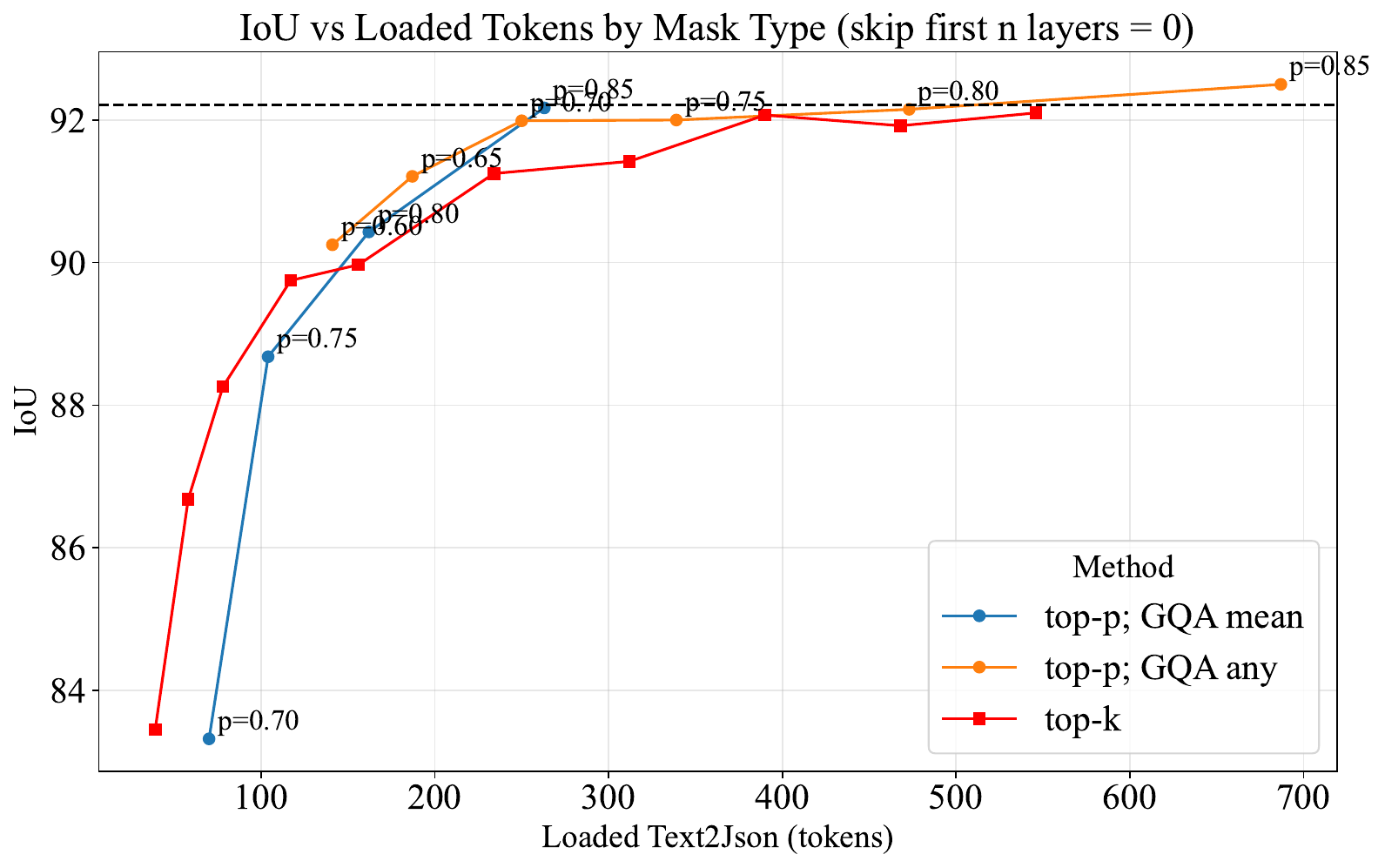}
    \caption{Trade-off between number of loaded tokens and generation quality for top-k and top-p methods. Results are shown for Llama-3.1-8B-Instruct on MultiNeedle (left), and for Qwen-3-4B-Instruct-2507 on Text2JSON (right).}
    \label{fig:appendix_top_p}
\end{figure}

In both the KV-cache offloading and sparse attention fields, there is growing interest in switching from conventional top-k token selection to promising top-p selection, which allows retrieving a dynamic number of tokens, potentially avoiding transfer for unnecessary tokens and utilizing more tokens upon request.  Papers such as Twilight ~\cite{lin2025twilightadaptiveattentionsparsity} and Progressive Sparse Attention (PSA) ~\cite{zhou2025progressivesparseattentionalgorithm} report improvements of their proposed top-p methods over top-k counterparts. Yet, under our evaluation protocol, switching from standard top-k to top-p showed little to no gain in terms of both generation quality and the number of transferred tokens. Figure~\ref{fig:appendix_top_p} summarizes our results. For grouped query attention (GQA)~\cite{lin2025twilightadaptiveattentionsparsity}, token selection is ambiguous because each key head is shared by multiple query heads and therefore receives multiple attention scores. We consider two ways to aggregate these scores. In ``GQA mean'', we average the attention scores within each query-head group and then apply top-$p$ selection. In ``GQA any'', we apply top-$p$ selection separately for each query head and load a token if it is selected by at least one query head in the group.

Another approach to dynamic token selection, extensively studied in KV-cache pruning---e.g., AdaKV~\cite{feng2024ada}, CAKE~\cite{qin2025cakecascadingadaptivekv}, and LAVa~\cite{shen2025lavalayerwisekvcache}---is to use a shared budget $K$ for all attention heads within a layer, rather than allocating a fixed budget of $K / \texttt{kv\_heads}$ to each head. Since different heads may require different numbers of tokens, reallocating the budget across heads can increase the total attention mass covered by the layer while keeping the overall token budget fixed. This can potentially improve generation quality without increasing the number of selected tokens. Having accurate attention scores during prefill, it is easy to determine which heads benefit more from receiving additional tokens. In pruning methods, such budget allocation is performed only once, during the prefill stage, because the remaining tokens are permanently evicted.

\begin{figure}[h]
    \centering
    \includegraphics[width=0.49\linewidth]{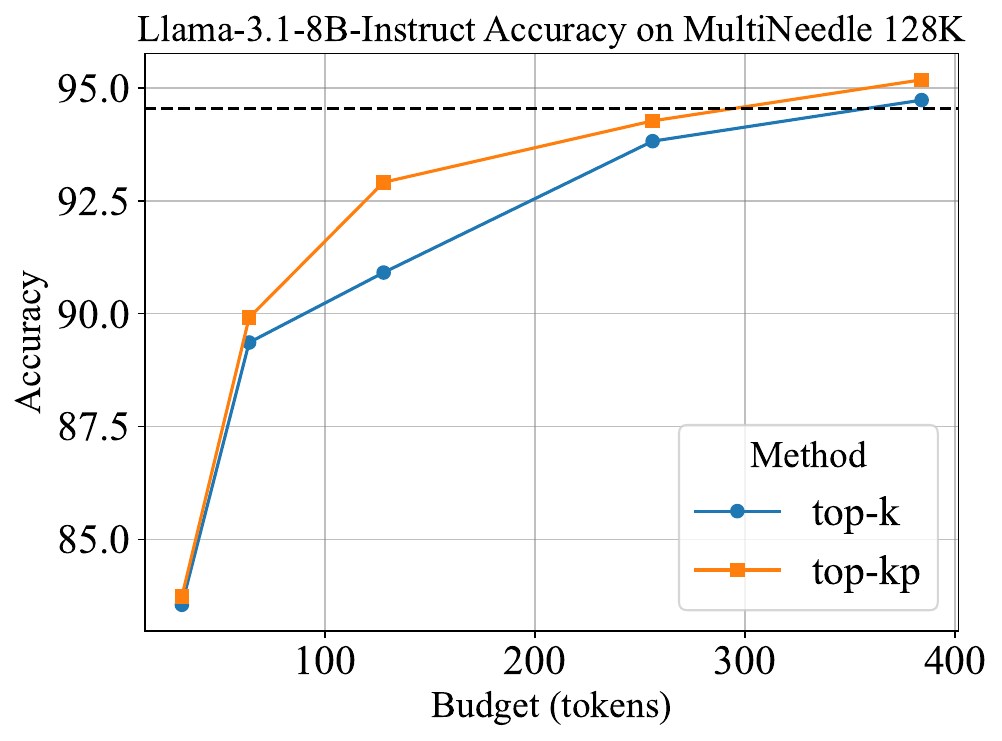}
    \includegraphics[width=0.465\linewidth]{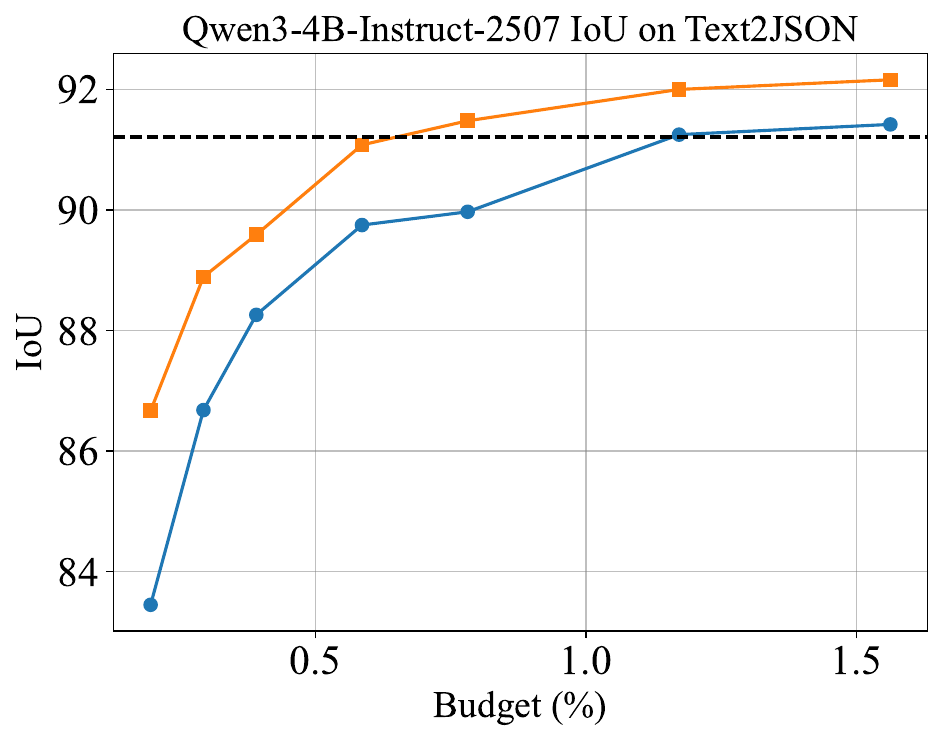}
    \caption{Generation quality for top-k and top-kp methods. Results are shown for Llama-3.1-8B-Instruct on MultiNeedle (left), and for Qwen-3-4B-Instruct-2507 on Text2JSON (right).}
    \label{fig:appendix_top_kp}
\end{figure}

In offloading, unlike pruning, a more optimal budget allocation could in principle be performed at every decoding step. However, this remains challenging because, unlike during prefill, offloading methods do not have access to ground-truth attention scores during decoding. Instead, their only source of information is the dot product between the current query and the ``landmarks``. To the best of our knowledge, dynamic budget allocation across heads has not been explored before in the context of KV-cache offloading. This may be partly because query--landmark dot products are unnormalized, making it difficult to compare token importance across heads. Moreover, as shown in Sections ~\ref{sect:experiments_budgets_groups} and ~\ref{sect:experiments_better_landmarks}, existing methods struggle to identify important tokens and often load unimportant ones due to key grouping. Since our proposed method excels at important tokens selection, we further experiment with dynamic head budgets. We find that this consistently improves generation quality under a fixed token budget; see Figure ~\ref{fig:appendix_top_kp}. We refer to this approach as top-$kp$: it uses the same fixed budget $K$ as top-$k$ methods, and performs selection based on attention scores rather than raw dot-product logits as in top-$p$ methods. Attention score is computed based on query and HIGGS 2-bit landmarks, and the $K$ tokens with the highest scores across all attention heads are selected.
However, due to the lack of a low-level implementation that supports attention computation with different heads attending to different numbers of tokens, we do not include this mechanism in the final YAKV method.

\section{Detailed Configurations for Section~\ref{sect:experiments_maintable}}\label{app:baseline_configs_and_extras}

We provide detailed configurations for KV offloading methods used in Section~\ref{sect:experiments_maintable}. We obtained these configurations as follows: we started with ShadowKV inference parameters from the original paper~\cite{sun2025shadowkv}, then constructed equivalent budgets for other algorithms by taking the official parameters and varying the offloading budget to equalize the amount of data loaded from RAM per forward:
\begin{itemize}
    \item \textbf{ShadowKV:} SVD rank 160, outlier budget 384, local budget 32, chunk size 8 (for landmarks), sparse budget = 1.5625\% of input length (loading Vs, keys on device).
    \item \textbf{InfiniGen:} alpha 99.0, capacity 1.0, partial weight ratio 0.3, budget = 0.78125\% of total KVs (see below). 
    \item \textbf{ArkVale:} page size 16 (cuboid-mean digest), sink tokens 32, window tokens = 64, unlimited layers = 0, pages loaded: 0.78125\% of total KV pages (rounded up).
    \item \textbf{LRQK:} rank = 32, recent = 64, budget = 0.78125\% of total KVs.
    \item \textbf{YAKV:} KV: 4.02 bit, Landmark: 2.02 bit, recent 64, sparse budget = 3.125\% input KVs.
\end{itemize}

For InfiniGen, the original implementation\footnote{\url{https://github.com/snu-comparch/InfiniGen}} does not support grouped query attention~\cite{ainslie2023gqa} --- we modify it to implement GQA-aware KV scoring using the same methodology as in ShadowKV. Note also that the original implementation can adaptively load less than the maximum number of KVs based on the score threshold (alpha). However, since our evaluations in Section~\ref{sect:experiments_maintable} use a fixed inference budget, it is advantageous to always load the maximum allowed number of KVs. For fair comparison, we set alpha=99, which enforces this behavior and slightly improves accuracy. 

However, unlike ShadowKV, InfiniGen stores original 16-bit precision keys in system memory (ShadowKV uses key SVD and only loads values). For this reason, InfiniGen loads half as many KVs to match the PCIe data transfer of ShadowKV.

Finally, original Infinigen implementation materializes full attention scores to select top-k keys for each query, which is not suitable for long context prefill. Thus we only perform selection during subsequent decoding steps to obtain upper bound on Infinigen accuracy.

For YAKV, we use standard 4- and 2-bit HIGGS~\cite{malinovskii2024pushing} grids as stated earlier: $d{=}2,n{=}256$ $d{=}4,n{=}256$ and adjust the budget accordingly.


\subsection{Dataset Sources, Versions and Licenses}\label{app:dataset_sources_versions_licenses}

\begin{itemize}
    \item \textbf{MultiNeedle:} using OpenCompass \url{https://github.com/open-compass/opencompass}, version 0.5.1.post1,  Apache-2.0 license.
    \item \textbf{Loong:} \url{https://github.com/mozerwang/loong}, commit ID \texttt{6d2115b},  Apache-2.0 license.
    \item \textbf{LongProc:} \url{https://github.com/princeton-pli/LongProc}, commit ID \texttt{673ec4c}, Apache-2.0 license. 
    \item Llama 3.x is under \url{https://www.llama.com/llama3/license/}.
    \item Qwen 3.x uses Apache-2.0 license, \url{https://huggingface.co/Qwen/Qwen3-30B-A3B-Instruct-2507/blob/main/LICENSE}.

\end{itemize}

\section{Alternative Quantization Experiments for YAKV}\label{app:yakv_extra_evals}

\begin{figure}[h]
    \centering
    \includegraphics[width=0.32\linewidth]{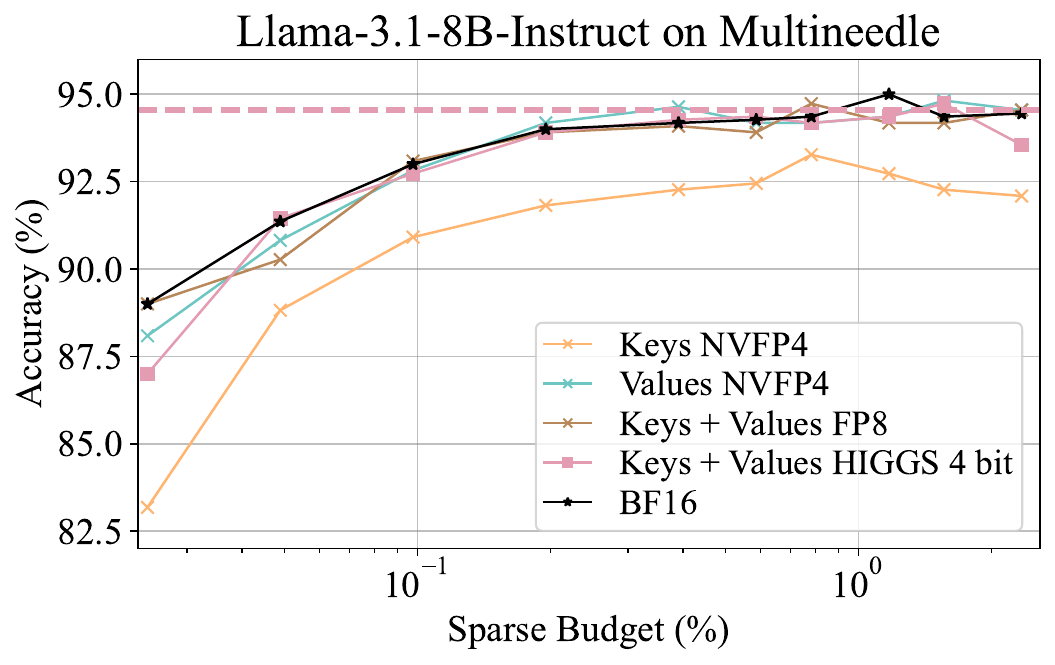}
    \includegraphics[width=0.32\linewidth]{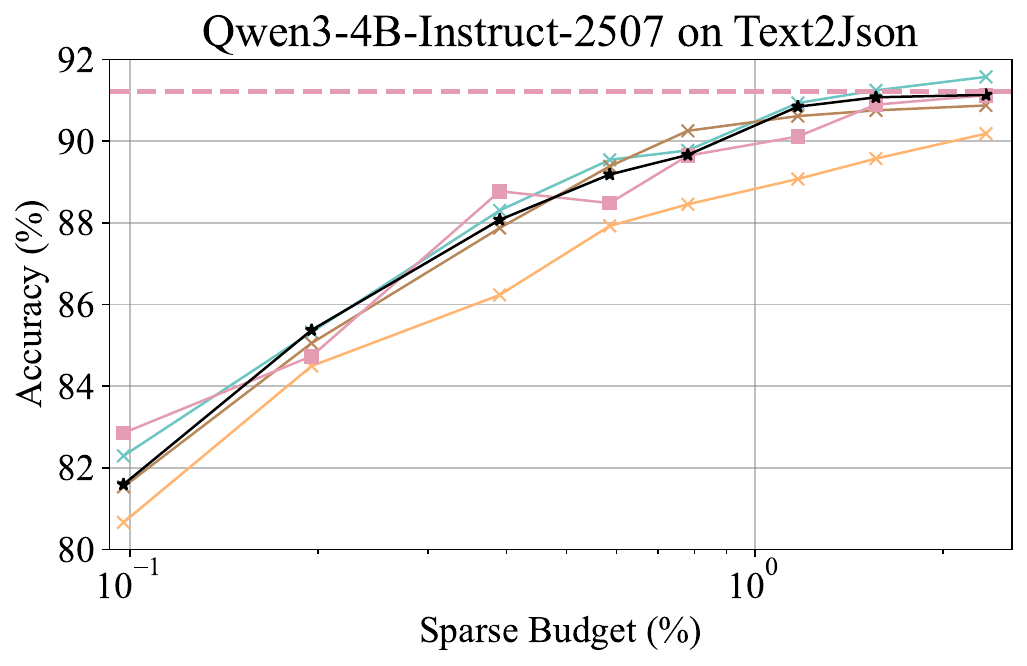}
    \includegraphics[width=0.32\linewidth]{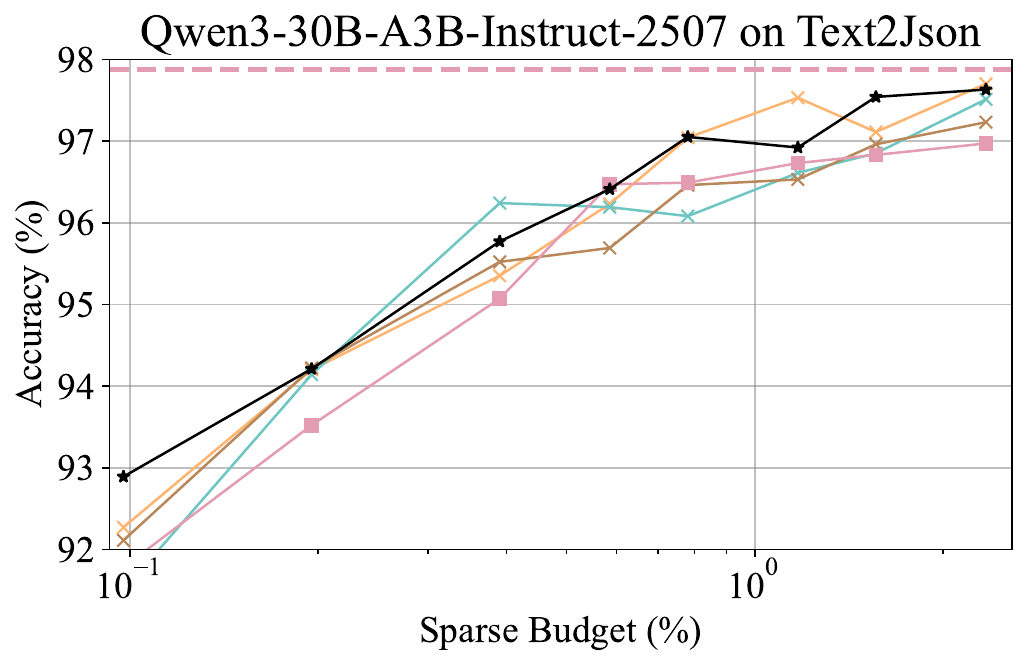}
    \caption{Comparison of different quantization methods. Results are shown for Llama-3.1-8B-Instruct on MultiNeedle (left), and for Qwen-3-4B-Instruct-2507 (middle) and Qwen3-30B-A3B-Instruct-2507 (right) on Text2JSON.}
    \label{fig:appendix_keys_values_quantization}
    \vspace{-15px}
\end{figure}

To avoid costly transfers between the CPU and GPU, different offloading methods propose various techniques for reducing the PCIe burden. For instance, ShadowKV ~\cite{sun2025shadowkv} retrieves only values from the CPU, while reconstructing keys from a compressed representation. InfiniGen ~\cite{lee2024infinigen} proposes overlapping the costly PCIe transfer of the keys and values for the next layer with the computation of the current layer. In contrast, we show that compressing both keys and values to as low as 4 bits reduces the number of bytes transferred from the CPU by a factor of 4, while maintaining generation quality on par with that obtained using bfloat16 keys and values. Before drawing this conclusion, we ablated various quantization methods for both keys and values, as shown in Figure~\ref{fig:appendix_keys_values_quantization}. Our results show that 4-bit quantization of keys and values onto the HIGGS grid preserves accuracy, whereas quantizing keys to NVFP4 may degrade quality.

\section{Inference Implementation Details for Section~\ref{sect:experiments_inference}}\label{app:inference_experiments}

We implement our inference code on top of Mini-SGLang \cite{minisglang}, a lightweight version of SGLang \cite{zheng2024sglang} designed for rapid research prototyping while retaining the high performance of the full-scale framework.
Unlike standard setups, we allocate the full KV cache in CPU memory.
In addition, we allocate the necessary workspace buffers, including an intermediate KV buffer for the selected KVs transferred from CPU memory to GPU memory.
This introduces minimal memory overhead because these buffers are allocated only once and shared across all model layers.
The intermediate KV buffer uses a contiguous layout rather than a paged layout, since paging provides no benefit in this context.
Since each key-value head has it's own set of tokens, we use per-head per-token quantization layout for HIGGS. This provides a substantial memory footprint from quantization scales, precisely $\frac{2}{head\_dim}$  from memory needed by full precision 16-bit KV-cache. We could reduce this overhead by concatenating each key-value entry into one vector or use a single scale for multiple consecutive tokens, but we implement neither of those to keep implementation as close as possible to described method. 

To support these operations, we develop several custom kernels.
The HIGGS quantization, scoring and CPU-to-GPU KV-copying kernels are implemented in CUDA, while the Top-K kernel is written in Triton. We fuse landmark dequantization and scoring kernels to effectively overlap computation and codebook lookup.
To accelerate memory transfers, we employ a memory-mapping strategy.
Instead of relying on explicit \texttt{cudaMemcpy} calls, we map the full CPU-resident KV tensor directly into the GPU address space, allowing GPU memory reads and writes to be translated into PCIe transactions.
This approach enables KV cache offloading without requiring substantial modifications to the core inference code.
Furthermore, it allows the GPU to fetch data without expensive per-layer CPU-GPU synchronization, which is a critical optimization because the tokens selected for loading are determined directly on the GPU.

In future work, we intend to further optimize the inference kernels.
We also plan to integrate widely used inference features, such as chunked prefill and a paged layout for the full KV cache.
Importantly, the current absence of these features does not compromise the validity or relevance of our performance benchmarks.


We measure inference speed using combined inputs from Text2JSON and MultiNeedle, with an average input length of approximately 38K tokens and a maximum input length of approximately 126K tokens.
With the bfloat16 Qwen3-30B-A3B model, the available VRAM is just sufficient to support batch size 1 in the full-attention baseline setup.
To simulate a real-world serving scenario, we continuously maintain the workload and keep the batch full.
We evaluate two setups: a prefill-heavy setting using the Instruct model and a decode-heavy setting using the Thinking model.
In Table~\ref{fig:app_4.5_detailed_perf}, we define RPS as responses per second, TTFT as time to first token, i.e., the time between sending a request and receiving the first token, and TPOT as time per output token.

\begin{table*}[h]
\centering
\small
\setlength{\tabcolsep}{6pt}
\renewcommand{\arraystretch}{1.2}
\caption{Detailed runtime performance evaluations on real data.}
\label{fig:app_4.5_detailed_perf}
\begin{tabular}{lcccccc}
\toprule
\textbf{Method} & \textbf{Batch} & \textbf{RPS} & \textbf{TTFT (ms)} & \textbf{TPOT (ms)} & \textbf{Output Length (tok)} & \textbf{Throughput (tok/s)} \\
\midrule
\multicolumn{7}{c}{\texttt{Qwen3-30B-A3B-Instruct-2507}} \\
\midrule
Baseline & 1 & 0.06 & 3100 & 38 & 359 & 26.3 \\
YAKV & 4 & 0.07 & 3319 & 124 & 349 & \textbf{32.3} \\
YAKV & 8 & 0.13 & 3401 & 163 & 367 & \textbf{49.1} \\
YAKV & 16 & 0.17 & 3828 & 247 & 366 & \textbf{64.8} \\
YAKV & 32 & 0.2 & 6164 & 420 & 371 & \textbf{76.2} \\
\midrule
\multicolumn{7}{c}{\texttt{Qwen3-30B-A3B-Thinking-2507}} \\
\midrule
Baseline & 1 & 0.011 & 3036 & 39 & 2289 & 25.6 \\
YAKV & 4 & 0.015 & 3985 & 108 & 2502 & \textbf{37.0} \\
YAKV & 8 & 0.026 & 6079 & 122 & 2560 & \textbf{65.6} \\
YAKV & 16 & 0.043 & 6659 & 150 & 2586 & \textbf{106.7} \\
YAKV & 32 & 0.044 & 10343 & 252 & 2381 & \textbf{127.0} \\
\midrule
\multicolumn{7}{c}{\texttt{Qwen3-32B}} \\
\midrule
Baseline & 1 & 0.02 & 9049 & 30 & 1416 & 33.3 \\
YAKV & 4 & 0.024 & 9088 & 97 & 1656 & \textbf{41.2} \\
YAKV & 8 & 0.033 & 9725 & 157 & 1539 & \textbf{51.0} \\
YAKV & 16 & 0.045 & 14341 & 214 & 1671 & \textbf{74.8} \\
\bottomrule
\end{tabular}
\end{table*}

\section{Hardware and Runtime Specifications}\label{app:hardware_runtime}

The specific hardware configurations necessary to reproduce our experiments vary by model size and experiment. However, all of the experiments reported in Sections~\ref{sect:experiments_svd},~\ref{sect:experiments_budgets_groups},~\ref{sect:experiments_better_landmarks},~and~\ref{sect:experiments_maintable} can be reproduced on a server with a single A100-80G GPU with 512 GiB RAM and 16 average performance (v)CPU cores. The runtime and throughput evaluations in Section~\ref{sect:experiments_inference} and Appendix~\ref{app:inference_experiments} require a single H200 GPU at PCIe gen. 5 x16 host-device interconnect and similar CPU / RAM requirements. They can be run with a different GPU (e.g. multiple weaker GPUs or a single B200), but the runtime and optimal batch sizes may be different.

In total, we estimate that our evaluations, preliminary experiments, development and re-runs (e.g. due to hardware failures) took ${\approx}8500$ A100-hours and fewer than 24 B200-hours. However, partial replication on a subset of models and/or benchmarks will take significantly less than that. Individual results for Qwen3-4B on Text2JSON, LongProc and MultiNeedle take 0.5-2 GPU-hours and can be run on 24-48GiB GPUs, depending on the KV offloading method.

Additionally, our evaluations on Loong use LLM-as-a-Judge verification. Individual evaluations in the canonical setting cost $\$10{-}30$, with total expenses within $\$1200$ across all main and preliminary experiments.

\section{Performance of hybrid architectures on context-intenstive tasks}\label{app:hybrid_models}

Recently, subquadratic alternatives to the attention mechanism have gained increasing popularity in LLM architectures. Replacing a large fraction of attention layers with modules such as Mamba \cite{gu2023mamba, dao2024transformersssms} or GatedDeltaNet \cite{yang2024gated} has been shown to achieve performance competitive with full-attention architectures while offering better scalability with increasing sequence length. However, the use of a fixed-size hidden state may potentially limit a model's ability to process and retain information over long contexts.

To investigate this hypothesis, we evaluated several models in which a subset—or all—of the attention layers were replaced with subquadratic alternatives. The results are presented in Table~\ref{tab:weak_hybrid_models} and Table~\ref{tab:qwen35_vs_qwen3}.

\vspace{-5px}
\begin{table}[h!]
\caption{Performance of Linear \& Hybrid attention models on context-intensive tasks.}
\label{tab:weak_hybrid_models}
\centering
\begin{tabular}{c c c }
\hline
\textbf{Model} & \textbf{MultiNeedle} & \textbf{Text2JSON} \\
\hline
Jamba-2-3B & 57.45 & 13.34 \\
LFM2.5-1.2B-Instruct & 0.00 & 3.68 \\
Olmo-Hybrid-Instruct-SFT-7B & 0.18 & 0.00 \\
\hline
\end{tabular}
\end{table}
\vspace{-5px}

\begin{table}[h!]
\caption{Comparison of Accuracy (\%) across different models and benchmarks. Average is the mean of the four Acc columns.}
\label{tab:qwen35_vs_qwen3}
\centering
\begin{tabular}{c c c c c c}
\hline
\textbf{Model} & \textbf{MultiNeedle} & \textbf{Text2JSON} & \textbf{Loong} & \textbf{LongProc} & \textbf{Average} \\
\hline
Qwen3-4B & 98.91 & 91.25 & 46.81 & 50.50 & 71.87 \\
Qwen3-30B-A3B & 100.00 & 97.88 & 56.29 & 59.70 & 78.47 \\
\hline
Qwen-3.5-4B & 99.91 & 83.98 & 56.60 & 46.14 & 71.66 \\
Qwen-3.5-9B & 100.00 & 96.10 & 60.61 & 55.67 & 78.10 \\
Qwen3.5-27B & 100.00 & 97.96 & 63.61 & 70.26 & 82.96 \\
Qwen3.5-35B-A3B & 100.00 & 98.02 & 60.92 & 68.34 & 81.82 \\
\hline
\end{tabular}
\end{table}

Notably, some hybrid architectures \cite{lfm2024report, groeneveld2024olmohybrid, lieber2024jamba} exhibit substantially degraded performance. Inspection of the generated responses revealed that most failures stem from either entering repetitive generation loops or violating the required output format. In contrast, the Qwen3.5 \cite{qwen35technicalreport} family achieves performance comparable to, and in some cases exceeding, that of the corresponding Qwen3 models employing full attention throughout the network.

These observations suggest that a relatively small number of self-attention layers may be sufficient to support context-intensive reasoning tasks, while the remaining layers can be replaced with more computationally efficient subquadratic alternatives without a significant loss in performance.

\vspace{-5px}
\section{Comparison of pre-RoPE and post-RoPE}\label{app:pqcache_and_magicpig}
\vspace{-5px}

Our main YAKV setup and inference experiments in Section~\ref{sect:experiments_inference} employ post-RoPE quantization to enable efficient inference by avoiding the need to rotate past keys at each step. However, some prior work on KV compression suggests that pre-RoPE quantization may yield higher accuracy.

In this section, we conduct additional experiments to compare post-RoPE YAKV with a pre-RoPE alternative. We evaluate YAKV using both pre- and post-RoPE quantization, keeping the same default hyperparameters as in Section~\ref{sect:experiments_maintable}. Table~\ref{tab:app_pre_post_rope} presents the comparison between pre- and post-RoPE YAKV. In the pre-RoPE variant, keys are quantized before RoPE and rotated after retrieval. Note that landmarks are always quantized post-RoPE to maintain inference efficiency.

The results indicate that pre-RoPE key quantization does not significantly improve model accuracy for YAKV. We attribute this to the randomized Hadamard transformation used in the HIGGS quantizer~\cite{malinovskii2024pushing}.

\begin{table}[htb]
\vspace{-5px}
\caption{Comparison of pre- and post-RoPE quantization in YAKV for Llama-3.1-8B-Instruct on MultiNeedle and Qwen3-4B-Instruct-2507 on Text2JSON.}
\vspace{5px}
\label{tab:app_pre_post_rope}
\centering
\begin{tabular}{c c c }
\hline
\textbf{Setup} & \textbf{Llama 3.1 8B MultiNeedle} & \textbf{Qwen3 4B Text2JSON} \\
\hline
Pre-RoPE & 93.36 & 90.83 \\
Post-RoPE & 93.73 & 91.05 \\
\hline
\end{tabular}
\end{table}


\end{document}